\documentclass[10pt,twocolumn,letterpaper]{article}

\usepackage[pagenumbers]{cvpr} 

\usepackage{tabularx, multirow}
\usepackage{adjustbox}
\usepackage[dvipsnames]{xcolor}

\definecolor{cvprblue}{rgb}{0.21,0.49,0.74}
\usepackage[pagebackref,breaklinks,colorlinks,citecolor=cvprblue]{hyperref}

\def\confName{CVPR}
\def\confYear{2024}

\title{WOUAF: Weight Modulation for User Attribution and Fingerprinting\\ in Text-to-Image Diffusion Models}

\author{Changhoon Kim\thanks{These authors contributed equally to this work.}$^{\;\:1}$ \quad Kyle Min\textsuperscript{\textasteriskcentered}$^{2}$ \quad Maitreya Patel$^1$ \quad Sheng Cheng$^1$ \quad Yezhou Yang$^1$ \\[1ex]
$^1$Arizona State University \qquad $^2$Intel Labs\\
{\tt\small \{kch,maitreya.patel,scheng53,yz.yang\}@asu.edu} \quad \tt\small kyle.min@intel.com
}

\DeclareUnicodeCharacter{03BB}{\ensuremath{\lambda}}

\begin{document}
\maketitle
\begin{abstract}
The rapid advancement of generative models, facilitating the creation of hyper-realistic images from textual descriptions, has concurrently escalated critical societal concerns such as misinformation. Although providing some mitigation, traditional fingerprinting mechanisms fall short in attributing responsibility for the malicious use of synthetic images. This paper introduces a novel approach to model fingerprinting that assigns responsibility for the generated images, thereby serving as a potential countermeasure to model misuse. Our method modifies generative models based on each user's unique digital fingerprint, imprinting a unique identifier onto the resultant content that can be traced back to the user. This approach, incorporating fine-tuning into Text-to-Image (T2I) tasks using the Stable Diffusion Model, demonstrates near-perfect attribution accuracy with a minimal impact on output quality. Through extensive evaluation, we show that our method outperforms baseline methods with an average improvement of 11\% in handling image post-processes. Our method presents a promising and novel avenue for accountable model distribution and responsible use. Our code is available in \url{https://github.com/kylemin/WOUAF}.

\end{abstract}    
\vspace{-0.1in}
\section{Introduction}
\label{sec:intro}

Recent advancements in generative models have propelled their proficiency, expanding their repertoire to include not just the generation of photorealistic images~\citep{karras2020analyzing,esser2021taming} but also the synthesis of images from textual prompts~\citep{nichol2022glide,saharia2022photorealistic,ramesh2022hierarchical, stable_diffusion}. These significant strides have equipped individuals with the capacity to leverage these models to create hyper-realistic images that correspond seamlessly with given textual instructions.

Nonetheless, the escalating prominence of generative models instigates pressing societal apprehensions. A case in point is Deepfake, intentionally crafted to disseminate misinformation, fostering a climate of fake news and political disarray~\citep{breland_2019, novak_2023, Lajka_2023}. 
The gravity of these concerns necessitates calls for governmental intervention to regulate the indiscriminate application of generative models\footnote{President Biden Issues Executive Order on Safe, Secure, and Trustworthy Artificial Intelligence.~\href{https://www.whitehouse.gov/briefing-room/statements-releases/2023/10/30/fact-sheet-president-biden-issues-executive-order-on-safe-secure-and-trustworthy-artificial-intelligence/}{The White House}}.

\begin{figure}[t]
  \centering
   \includegraphics[width=0.8\linewidth]{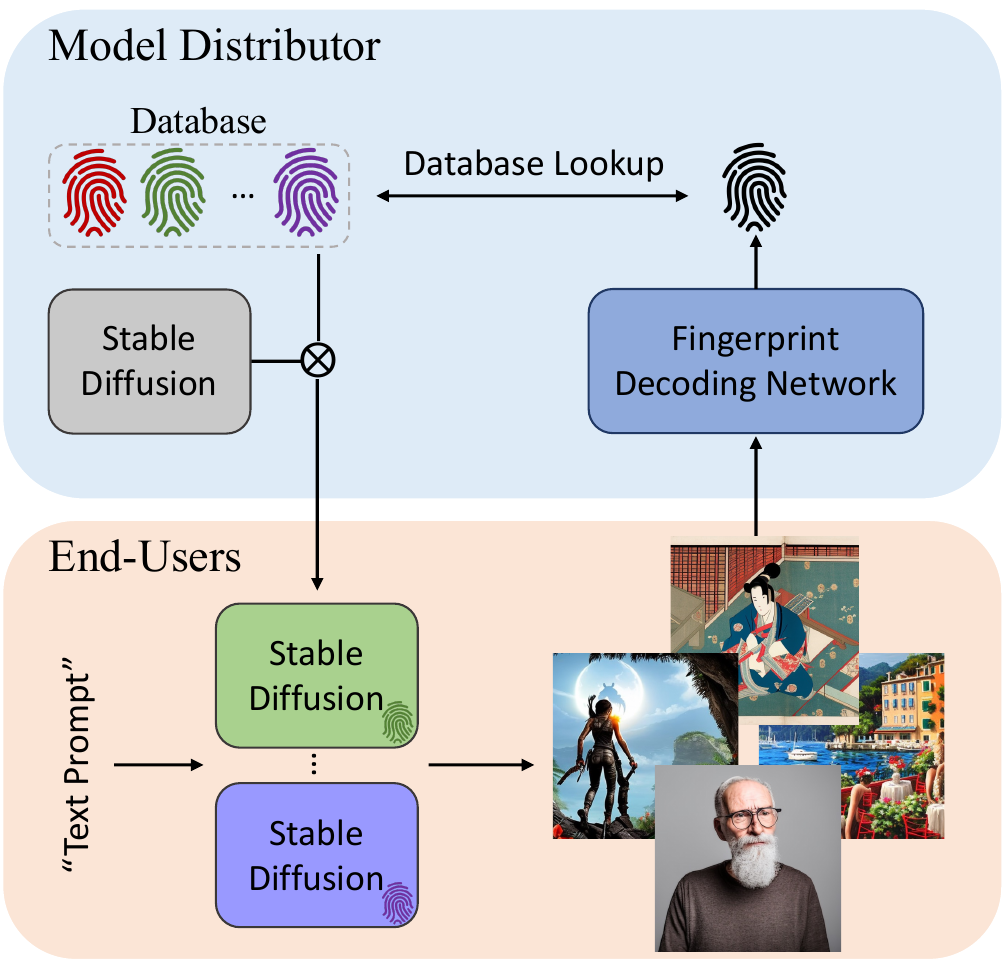}
   \caption{Illustration of user attribution based on our method. Please refer to the main text for detailed descriptions.}
   \label{fig:fig1}
   \vspace{-0.15in}
\end{figure}

A feasible method to counteract malicious use involves assigning accountability for generated images. One approach to achieve this is by integrating independent fingerprinting modules that can embed user-specific information on top of image generation. The open-source Text-to-Image (T2I) project Stable Diffusion (SD)~\citep{stable_diffusion} currently employs this technique using discrete wavelet transform or RivaGAN~\citep{rivagan}. However, in the open-source setting, bypassing the fingerprinting module is straightforward and can be achieved by commeting just a single line in the source code~\cite{stable_signature}.

\textit{Is it feasible to achieve user attribution without an independent fingerprinting module?}
In response, we propose a distributor-oriented methodology named \textbf{WOUAF}, standing for \textbf{W}eight m\textbf{O}dulation for \textbf{U}ser \textbf{A}ttribution and \textbf{F}ingerprinting.
In practical terms, when a model inventor open-sources their work to a model distributor such as Huggingface, the distributor could utilize our proposed method to create a generic version. Upon receiving a download request from an end-user, the distributor can adjust the model weights using our technique and deploy a fingerprinted version to the user, simultaneously registering the user's fingerprint into their database. In the event of a model's malicious exploitation, the distributor can decode the fingerprint from the misused image and cross-reference it with their database to identify the responsible user. Consequently, this provides the distributor with an actionable method to counteract malicious uses of the model (see \cref{fig:fig1} for a comprehensive framework of our methodology).

Our methodology, designed for T2I tasks, is integrated into the Stable Diffusion (SD) framework without necessitating any structural changes to the model. This design choice effectively prevents end-users from bypassing the fingerprinting process. Consistent with prior research~\cite{kim2020decentralized,yu2020responsible,yu2021artificial,nie2023attributing,stable_signature}, our primary goal is to maintain high attribution accuracy while ensuring minimal impact on output quality, as elaborated in~\cref{sec:method}. Our rigorous evaluations of this method concentrate initially on assessing both attribution accuracy and image quality. We have found that our approach attains nearly flawless attribution accuracy with only a slight influence on image quality. Moreover, we evaluate the robustness of our method in various scenarios involving post-processing manipulations that images might undergo. Our method outperforms baseline methods in these robustness tests, showing an average improvement of 11\% in handling such manipulations (refer to~\cref{sec:experiments} for further details).

There are four main contributions: 
(1) We introduce \textbf{WOUAF}, a distinctive distributor-centered fine-tuning methodology. This approach embeds fingerprints within the model in such a way that end-users cannot easily circumvent or remove them.
(2) Our method successfully achieves high attribution accuracy, while maintaining the quality of the output images.
(3) Our approach exhibits marked resilience against a diverse array of image post-processes, a vital attribute for practical applications.
(4) We conduct thorough assessments to balance attribution accuracy with manipulations to intentionally remove fingerprints, including strategies like image compression via auto-encoders and obfuscation through model fine-tuning.

\section{Related Work} \label{sec:related_work}

In this section, we discuss related works of model fingerprinting in generative models. More related works are available in the appendix.

\medskip\noindent\textbf{Inventor-oriented Model Fingerprinting.}
Yu et al.~\citep{yu2021artificial} leveraged a pre-trained deep steganography model to embed fingerprints into the training set for fingerprinted GANs. However, this approach suffers from limited scalability, as it necessitates training a GAN from scratch for each distinct fingerprint. To address this issue, Yu et al.~\citep{yu2020responsible} introduced a weight modulation method~\citep{karras2019analyzing} that directly embeds a user's fingerprint into the generator's weights.
Despite these advancements, current methods are predominantly tailored for GAN-based models and typically require training from scratch. This raises important questions regarding their suitability for diffusion-based models, which have a different structural makeup compared to GANs, and the feasibility of avoiding the requirement for training from scratch. The adoption of fine-tuning as a method for embedding fingerprints presents a promising solution. It facilitates the incorporation of fingerprints into pre-trained diffusion models, eliminating the necessity for comprehensive retraining from the ground up~\cite{stable_signature, receipe_for_wm_diffusion}. This approach significantly streamlines the process, allowing model inventors to concentrate on core model development without the complexities of embedding fingerprints during training.

\medskip\noindent\textbf{Distributor-oriented Model Fingerprinting.}
Kim et al.~\cite{kim2020decentralized} proposed a technique for achieving user attribution by explicitly incorporating user-specific fingerprints into the generator’s output.
While this simplified attribution method allowed for the derivation of sufficient fingerprint conditions, it necessitates a trade-off between the quality of the generated output and attribution accuracy, which is further exacerbated when image post-processes are taken into account. 
To tackle this issue, an approach has been proposed that utilizes subtle semantic variations along latent dimensions as fingerprints, generated by perturbations of eigenvectors in the latent distribution~\cite{nie2023attributing}.
This method demonstrates an improved balance between generation quality and attribution accuracy. However, its applicability is restricted to unconditional image generation, as eigenvectors are computed by sampling the learned latent representation. In the context of conditional image generation, estimating eigenvectors of latent representation becomes challenging due to the vast space of conditions, such as those found in text conditions.

\begin{figure*}[ht]
\centering
  \includegraphics[width=\textwidth]{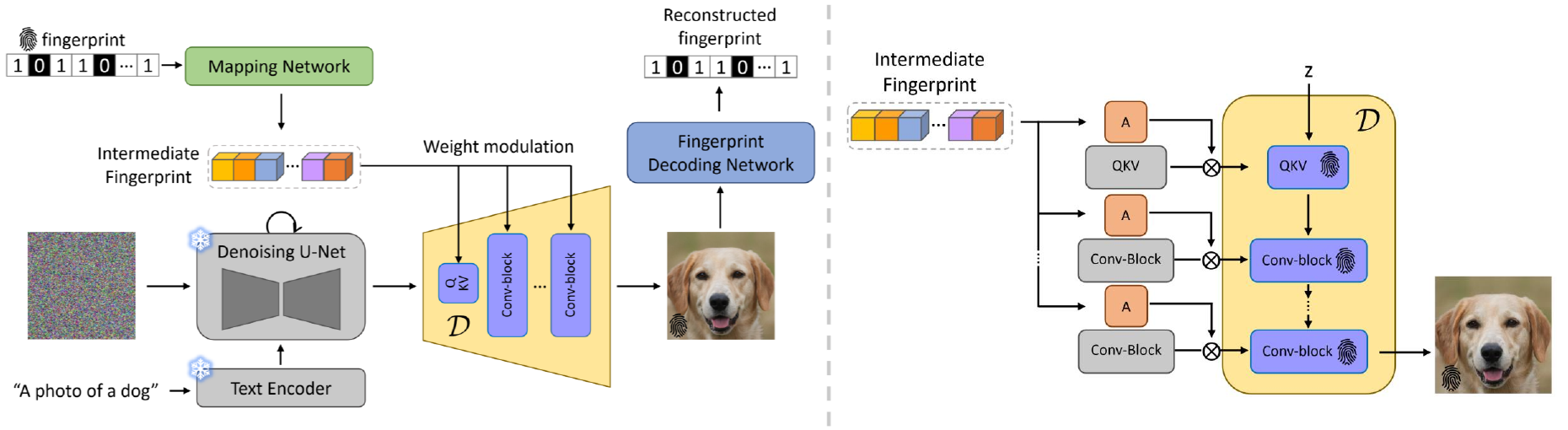} \\[-0.1ex] \hspace*{2em}(a) The overall pipeline. \hspace{15em}(b) Weight modulation.
  \caption{Depiction of our method's pipeline and weight modulation: 
  (a) The model fingerprinting procedure encompasses encoding via the mapping network and weight modulation, along with decoding through the fingerprint decoding network.
  (b) Weight modulation of the decoding network $\mathcal{D}$ to incorporate the fingerprint.}
  \label{fig:overview}
  \vspace{-0.15in}
\end{figure*}

\medskip\noindent\textbf{Recent Advances in Fingerprinting for Text-to-Image Diffusion Models.}
Recent studies~\cite{receipe_for_wm_diffusion,stable_signature,tree-ring} have scrutinized fingerprinting techniques in the Stable Diffusion model~\cite{stable_diffusion}, uncovering vulnerabilities in existing methods~\cite{rivagan} that facilitate easy circumvention~\cite{stable_signature} or robust post-hoc fingerprinting~\cite{tree-ring}. Fernandez et al.~\cite{stable_signature} achieved near-perfect attribution accuracy by fine-tuning user-specific models to align with steganography module~\cite{zhu2018hidden}, demonstrating a viable alternative to conventional post-hoc fingerprinting modules~\cite{rivagan}. However, this approach scales linearly in computational demand with the number of users since it necessitates fixed-time fine-tuning for each individual. In contrast, our method requires only a one-time training followed by a negligible forward pass time to generate user-specific models. Furthermore, our approach shows superior robustness against common image post-processing techniques compared to that of Stable Signature~\cite{stable_signature} (refer to Section~\cref{subsec:robustness-postprocesses} for details).

Another notable contribution is by Wen et al.~\cite{tree-ring}, who introduced an alternative fingerprinting method for the Stable Diffusion model~\cite{stable_diffusion}. Their method, similar to post-hoc style fingerprinting~\cite{rivagan}, depends on user-driven embedding, which allows end-users the option to exclude the fingerprint. Moreover, it is confined to the DDIM scheduler~\cite{ddim}. Our method, in contrast, is adaptable to both the DDIM~\cite{ddim} and Euler schedulers~\cite{euler}, underscoring its versatility and wider applicability (refer to the Appendix).

\section{Methods}\label{sec:method}

This section outlines our approach, beginning with an overview of the Text-to-Image (T2I) diffusion model with a focus on the Stable Diffusion (SD) model~\cite{stable_diffusion} detailed in~\cref{subsec:stable_diffusion}. We then introduce our key component, the user-specific weight modulation, in~\cref{subsec:weight_modulation}. The section concludes with a detailed explanation of our training objectives and methods, outlined in~\cref{subsec:training_objective}.

\begin{table*}[t]
\centering
\caption{Evaluation of Attribution Accuracy and Image Generation Quality. We conducted validation using the MS-COCO~\cite{lin2014microsoft} test set and the LAION-Aesthetics~\cite{schuhmann2022laion} dataset, which were excluded from our training phase. Symbols $\uparrow$ and $\downarrow$ denote preferred higher and lower values, respectively.}
\label{tab:attribution_acc}
{\small
\begin{tabularx}{\textwidth}{lXXXXXXX}
\toprule
\multirow{3}{*}{Model} & &\multicolumn{3}{c}{MS-COCO} & \multicolumn{3}{c}{LAION} \\\cmidrule(lr){3-5} \cmidrule(lr){6-8}
 &Fingerprinting Time ($\downarrow$) & Attribution Acc ($\uparrow$) & CLIP-score ($\uparrow$)& FID ($\downarrow$)
 & Attribution Acc ($\uparrow$) & CLIP-score ($\uparrow$)& FID ($\downarrow$)\\
\midrule
Original SD~\citep{stable_diffusion} & -  & - & 0.73 & 24.48 & - & 0.50 & 19.67 \\
DAG~\cite{kim2020decentralized} & 8.4 hr & 0.70 & 0.73 & 26.54 & 0.71 & 0.49 & 23.13 \\
Stable Signature~\cite{stable_signature} & $<$ 1 min & 0.99 & 0.73 & 24.55 & 0.98 & 0.50 & 20.02\\
\midrule
WOUAF-conv & $<$ 1 sec & 0.99 & 0.73 & 24.43 & 0.98 & 0.51 & 20.46 \\
WOUAF-all & \textbf{$<$ 1 sec} & \textbf{0.99} & \textbf{0.73} & \textbf{24.42} & \textbf{0.99} & \textbf{0.51} & \textbf{19.91} \\
\bottomrule
\end{tabularx}
}
\vspace{-0.1in}
\end{table*}

\subsection{Preliminaries} \label{subsec:stable_diffusion}

Our approach utilizes the Stable Diffusion (SD) model, which functions within the latent space framework of an autoencoder.
SD comprises two main elements: Firstly, an autoencoder is pre-trained on an extensive dataset of images. Its encoder, $\mathcal{E}(\cdot):\mathbb{R}^{d_x} \rightarrow \mathbb{R}^{d_z}$, converts an image $x \sim p_{data}$ into a latent representation $z = \mathcal{E}(x)$. The decoder, $\mathcal{D}(\cdot):\mathbb{R}^{d_z} \rightarrow \mathbb{R}^{d_x}$, then reconstructs the original image from this latent representation, resulting in $\hat{x} = \mathcal{D}(z)$.
The secondary element is a diffusion model, based on the U-Net architecture~\cite{ronneberger2015u}, represented as $\epsilon_{\theta}$. This model is adept at generating latent representations and can be conditioned using pre-trained text embeddings.

\subsection{User-specific Weight Modulation} \label{subsec:weight_modulation}
Our method is fundamentally based on integrating fingerprints into the parameters of the SD through weight modulation~\citep{karras2019analyzing, yu2020responsible}.

The overall pipeline of our method is illustrated in Fig.~\ref{fig:overview}(a). A user-specific fingerprint is drawn from a Bernoulli distribution with a probability of 0.5, represented as $\phi \in \Phi := \text{Bernoulli}(0.5)^{d_\phi}$, where $d_\phi$ signifies the fingerprint length in bits. We employ a mapping network $\mathcal{M}(\cdot):\mathbb{R}^{d_\phi} \rightarrow \mathbb{R}^{d_M}$ to convert the sampled fingerprint $\phi$ into an intermediate fingerprint representation within the $d_M$ dimension. For modulating each layer in the SD component, we introduce an affine transformation layer, $\mathcal{A}_l(\cdot):\mathbb{R}^{d_M} \rightarrow \mathbb{R}^{d_j}$, for all layers $l$. As depicted in \cref{fig:overview}(b), this transformation matches the dimensions between $d_M$ and the $j$-th channel in weight $W \in \mathbb{R}^{i,j,k}$, where $i,j,k$ denote input, output, and kernel dimensions, respectively. The weight modulation for the $l$-th layer is defined as:
\begin{equation}
    W_{i,j,k}^\phi = u_{j} * W_{i,j,k},
\end{equation}
where $W$ and $W^\phi$ denote the pre-trained and fingerprinted weights respectively, $u_j = \mathcal{A}_l(\mathcal{M}(\phi))$ is the scale of the fingerprint representation corresponding to the $j$th output channel. 

We incorporate fingerprints into the SD by applying weight modulation exclusively to the weights in the decoder $\mathcal{D}$. The rationale for not applying modulation to both the diffusion model $\epsilon_\theta$ and decoder $\mathcal{D}$, an approach that mirrors GAN-based models~\citep{yu2020responsible}, is explained in~\cref{sec:benefit_of_only_decoder}.

\subsection{Training Objectives} \label{subsec:training_objective}
Our training architecture comprises two primary objectives. The initial objective is to decode fingerprints from the provided images. We train a fingerprint decoding network $\mathcal{F}(\cdot):\mathbb{R}^{d_x} \rightarrow \mathbb{R}^{d_\phi}$, which is instantiated by ResNet-50~\citep{resnet}, as follows:
\begin{align}\label{eq:recon}
   L_{\phi} = \mathbb{E}_{z=\mathcal{E}(x), \phi \sim \Phi}\sum_{i=1}^{d_\phi} [\phi_i &\log \sigma(\mathcal{F}(\mathcal{D}(\phi, z))_i \nonumber \\
  +  (1-\phi_i)&\log(1-\sigma(\mathcal{F}(\mathcal{D}(\phi, z)))_i ], 
\end{align}
where $\sigma(\cdot)$ refers to the sigmoid activation function, constraining the output of $\mathcal{F}$ to the range $[0,1]$. Thus, this loss function effectively combines binary cross-entropy for all bits of the fingerprint.
During training time, fingerprint $\phi$ is sampled from Bernoulli distribution.
However, after training, the model distributor initially samples a user-specific fingerprint $\phi_\alpha$ and subsequently modulates the decoder $\mathcal{D}$ using $\phi_\alpha$. The user will receive the fingerprinted decoder $\mathcal{D}(\phi_\alpha, \cdot)$, which solely permits latent input.

The secondary objective endeavors to regularize the quality of outputs. Ideally, this regularization inhibits the decoder $\mathcal{D}$ from compromising image quality while minimizing $L_{\phi}$ in~\cref{eq:recon}:
\begin{equation}\label{eq:quality}
    L_{\text{quality}} = \mathbb{E}_{z=\mathcal{E}(x), \phi \sim \Phi} \left[\ell(x, \mathcal{D}(\phi, z))\right],
\end{equation}
$\ell$ represents the distance metric between original images and fingerprinted images.
For practical applications, we utilize perceptual distance~\citep{lpips} to gauge the perceptual difference between $x$ and $\mathcal{D}(\phi,z)$.

The final objective function can be formulated as:
\begin{equation}\label{eq:final_loss}
    \min_{\mathcal{A},\mathcal{M},\mathcal{D},\mathcal{F}} \lambda_1 L_{\phi} + \lambda_2 L_{\text{quality}},
\end{equation}
where both $\lambda_1$ and $\lambda_2$ are set to $1.0$. Fundamentally, the loss function aspires to reconstruct fingerprints while maintaining the quality of the generated outputs. To assess the efficacy of our proposed method, we employ attribution accuracy and image quality metrics (Refer to~\cref{sec:evaluation} for details).

\newcommand{\mathbbm}[1]{\text{\usefont{U}{bbm}{m}{n}#1}}
\section{Experiments} \label{sec:experiments}

\subsection{Experiment Settings}
\noindent\textbf{Datasets.}
Our approach is fine-tuned on the MS-COCO~\cite{lin2014microsoft} dataset, adopting the Karpathy split. For methodological evaluation, we harness the test set from MS-COCO and randomly sample from the LAION-aesthetics~\citep{schuhmann2022laion} dataset. 
For T2I image generation, we adopt the Euler scheduler~\citep{euler} with timestep $T=20$, and the classifier-free guidance scale~\citep{ho2022classifier} is set to 7.5 unless otherwise specified. Evaluation for DDIM scheduler~\cite{ddim} and various image generation hyperparameters are available in the Appendix.

\begin{figure*}[t]
\centering
  \includegraphics[width=\textwidth]{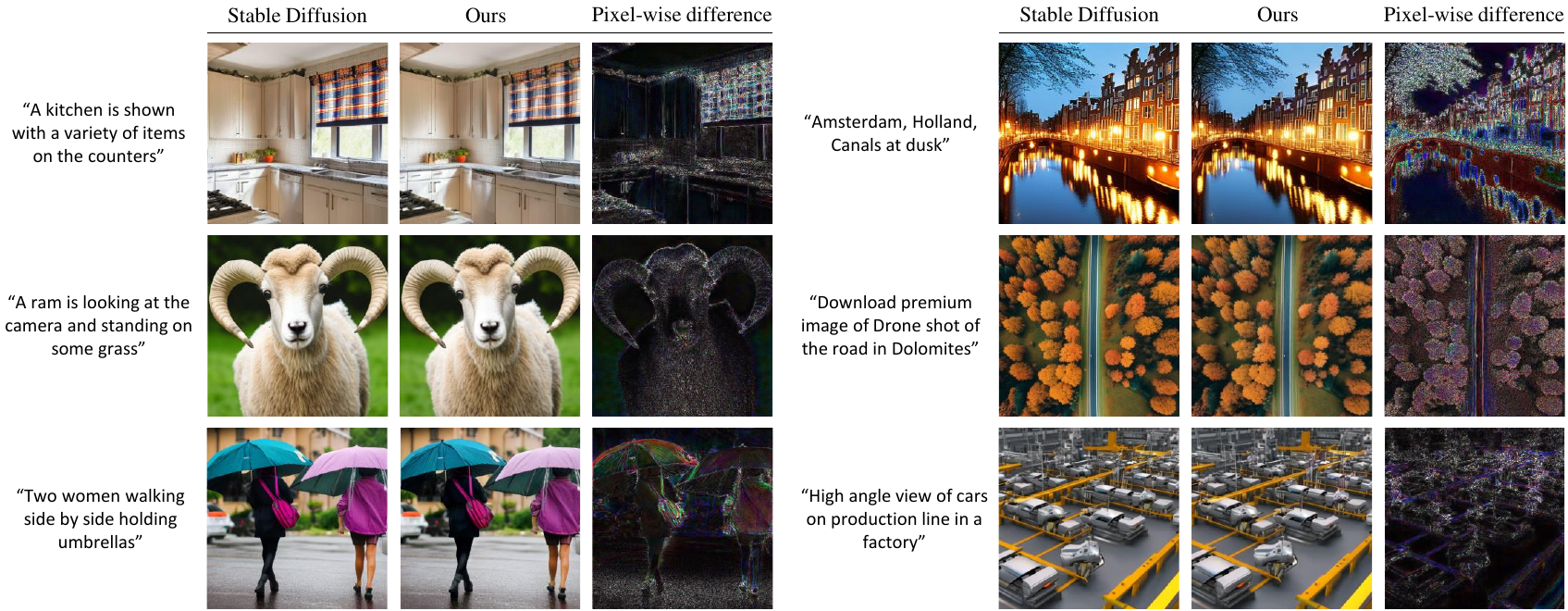} \\[-0.1ex] \hspace*{7.5em}(a) MS COCO \hspace{17.0em}(b) LAION Aesthetics
  \caption{Qualitative comparison of the original and fingerprinted Stable Diffusion models on MS-COCO~\cite{lin2014microsoft} and LAION aesthetics~\cite{schuhmann2022laion} (Pixel-wise differences$\times$ 5: they are multiplied by a factor of 5 for better view). We can observe that our method maintains high image quality.}
  \label{fig:qualatative_example}
  \vspace{-0.15in}
\end{figure*}

\medskip\noindent\textbf{Experimental Setting.}
We implement the weight modulation following the design specified in the source code of StyleGAN2-ADA~\cite{karras2020training}. Our mapping network $M$ is designed with a series of fully connected layers, wherein all experiments are conducted using a two-layer configuration. 
To train robust models against image post-processing transformations, differentiable post-processes are necessary. To this end, we incorporate the Kornia library~\cite{eriba2019kornia}. For Stable Signature~\cite{stable_signature}, we utilize the official code provided by the authors. We note that its post-processing transformations are replaced with our version for fair comparison. Appendix includes details on mapping network dimensions, training parameters, and optimizer.

\medskip\noindent\textbf{Evaluations.}\label{sec:evaluation}
User attribution accuracy is gauged by the formula:
$\frac{1}{d_\phi} \sum_{i=1}^{d_\phi} \mathbbm{1}(\phi_i = \hat{\phi}_i)$,
where $\phi$ is the true fingerprint and $\hat{\phi} = \mathbbm{1}\left[\sigma(\mathcal{F}(x_{\phi})) > 0.5\right]$ is the estimated fingerprint from image $x_{\phi}$. Unless otherwise stated, $d_\phi$ is set to 32 in our experiments (Refer to~\cref{sec:capacity} for additional information). We further employ a statistical test~\cite{yu2021artificial, yu2020responsible} to evaluate matching bits between $\hat{\phi}$ and $\phi$.
The null hypothesis $H_0$ suggests that the number of matching bits arises by chance. The test uses a binomial distribution, with a $p$-value derived as: $P(X \geq k | H_0) = \sum^{d_\phi}_{i=k} {{d_\phi}\choose{i}} 0.5^{d_\phi}$. A $p$-value below 0.05 leads to the rejection of $H_0$, with $1-p$ serving as an indicator of verification confidence.
Lastly, to validate the quality of our method, we assess image quality using the Fréchet Inception Distance (FID)~\citep{heusel2017gans} and employ the Clip-score~\citep{hessel2021clipscore} to determine the alignment between text and generated images. Additional experimental details can be found in the Appendix.

\medskip\noindent\textbf{Models.}
For evaluating our methodology, we benchmark against two established baseline methods: DAG~\cite{kim2020decentralized} and Stable Signature~\cite{stable_signature}. Both these methods, conceptualized from the model distributor's standpoint, incorporate fine-tuning for model fingerprinting. To ensure a fair comparison, we retrain the Stable Signature method within our training settings by replacing its post-processing scheme.
Additionally, we evaluate our method against three distinct variants based on the specific layers chosen for weight modulation implementation. The first variant, WOUAF-conv, applies modulation to only the convolutional layers in $\mathcal{D}$. In contrast, WOUAF-all extends this approach across all layers of $\mathcal{D}$, covering both self-attention and convolution layers. The final variant implements weight modulation in both the diffusion model $\epsilon_\theta$ and the decoder $\mathcal{D}$, mirroring the approach used in GAN-based methods~\citep{yu2020responsible}. Further details on why this variant is not used in our experiments are discussed in~\cref{sec:benefit_of_only_decoder}.

\subsection{Fingerprint Capacity}\label{sec:capacity}

\begin{table}[t]
\caption{Experiments of attribution accuracy across various fingerprint dimensions ($d_{\phi}$).}
\centering
\begin{tabular}{lcccc}
\toprule
& \multicolumn{4}{c}{Attribution Accuracy} \\
\cmidrule(r){2-5}
Fingerprint Dims. & 16 & 32 & 64 & 128 \\
\midrule
WOUAF-conv & 0.99 & 0.99 & 0.98 & 0.94 \\
WOUAF-all & 0.99 & 0.99 & 0.99 & 0.97 \\
\bottomrule
\end{tabular}
\label{tab:key_capacity}
\vspace{-0.1in}
\end{table}

The capacity of our method depends on the maximum number of unique user-specific fingerprints it can support without significant crosstalk. This capacity is primarily influenced by the fingerprint dimension ($d_\phi$). Selecting an optimal $d_\phi$ presents a challenge: while a larger $d_\phi$ can accommodate more users, it also complicates effective fingerprint decoding~\cite{Li2021ASO}.

To investigate this trade-off, we conduct an analysis with varying fingerprint dimensions, specifically $d_\phi$ values of 16, 32, 64, and 128. \cref{tab:key_capacity} presents the user attribution accuracy for each $d_\phi$ value. As shown in~\cref{tab:key_capacity}, attribution accuracy tends to decrease monotonically as $d_\phi$ increases.
Importantly, both our variant models achieve a near-perfect attribution accuracy of 0.99 for $d_\phi$ values of 16, 32, and 64. However, for $d_\phi = 128$, WOUAF-all variant outperforms the WOUAF-conv variant. For a balanced comparison with existing methods, we choose $d_\phi=32$, which notably can support a substantial user base exceeding 4 billion $\approx 2^{32}$.

\subsection{Attribution Accuracy and Image Quality}
\label{sec:att_and_quality}
We conduct a comprehensive evaluation of WOUAF, focusing on attribution accuracy and image quality. The assessment involves the MS-COCO~\cite{lin2014microsoft} test set and the LAION-Aesthetics~\cite{schuhmann2022laion} dataset, which are excluded from the training phase. The results, detailed in~\cref{tab:attribution_acc}, showcase the efficacy of our method.

Our variants, namely WOUAF-conv and WOUAF-all, demonstrate superior performance in attribution accuracy over DAG~\cite{kim2020decentralized}, indicating their proficiency in accurately decoding embedded fingerprints from the generated images. These variants also show competitive results when compared to Stable Signature~\cite{stable_signature}, reinforcing our methodology's robustness. Notably, we achieve this high level of accuracy without significantly compromising image quality. Both FID scores and Clip-scores showed minimal variation from the baseline SD model, indicating that our approach has a negligible impact on image output quality. This is further corroborated by qualitative examples in~\cref{fig:qualatative_example}, which highlight WOUAF's ability to reliably incorporate fingerprinting without degrading image generation quality. For additional insights, uncurated image collections are provided in the Appendix.

Given the growing importance of T2I models, computation time for fingerprinting emerges as a key metric. Our method stands out in computational efficiency. It contrasts with approaches like Stable Signature that need fine-tuning for each new fingerprint. Our method requires just a single forward pass, markedly reducing computational overhead.

\subsection{Attribution Analysis for Diverse Image Sources}\label{subsec:attribution_diverse_sources}
Investigating the attribution of generated images to responsible users, we explore the potential for images from non-fingerprinted or varied sources to bypass our system. Our analysis aims to determine if decoded fingerprints from such images match any entries in the model distributor's database. A mismatch indicates the image's external origin, absolving users in the database.

We adopt the experimental setup from~\cite{yu2020responsible}, compiling a dataset with different image types: authentic images from the MS-COCO test set~\cite{lin2014microsoft}, non-fingerprinted images from Stable Diffusion~\cite{stable_diffusion}, and synthesized images from ProGAN~\cite{karras2017progressive}, StyleGAN~\cite{karras2019style}, and StyleGAN2~\cite{karras2019analyzing}, with each category containing 1,000 samples. Given our extensive user database of 1 million entries, we set a threshold at $32 * 0.95 \approx 30$ bits, aligning with our 0.99 attribution accuracy as shown in~\cref{tab:attribution_acc}.

Our rigorous experiments revealed that, irrespective of the source, no images were incorrectly attributed as possessing a fingerprint from our 1 million fingerprint database. This reinforces the reliability of our attribution approach as detailed in~\cref{sec:att_and_quality} demonstrating the robustness of our system against diverse image sources.

\begin{figure}[t]
\centering
  \includegraphics[width=\linewidth]{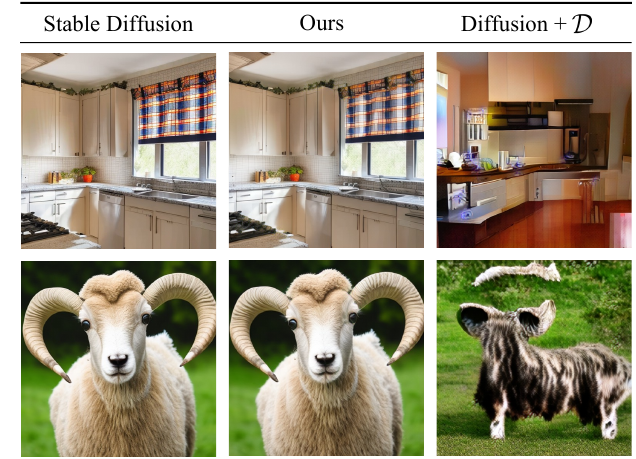}
  \caption{Comparative analysis of weight modulation on decoder $\mathcal{D}$ and diffusion model $\epsilon_{\theta}$ with decoder $\mathcal{D}$. Modulating the diffusion model negatively affects image quality. 
  }
  \label{fig:benefits_only_decoder}
  \vspace{-0.1in}
\end{figure}

\subsection{Benefits of Finetuning only Decoder}
\label{sec:benefit_of_only_decoder}
When developing our last variant that incorporates weight modulation into both the diffusion model $\epsilon_\theta$ and the decoder $\mathcal{D}$, we note that the resultant pipeline demonstrates similarities with the GAN-based method~\citep{yu2020responsible}. A direct comparison between ours and the GAN-based methods may not be entirely straightforward, given the fundamental differences in their training methodologies. This is because the GAN-based methods entail training from scratch, whereas our proposed approach leans towards fine-tuning. Nevertheless, both methodologies share a common mechanism: they aim to modulate the weights of the layers instrumental in learning the latent space. The shared characteristic underscores the fundamental objective of optimizing the balance between attribution accuracy and generation quality.

However, our empirical observations suggest that this variant
does not consistently achieve commendable performance as an attribution model. Specifically, it appears that this variant can only optimize either attribution accuracy or generation quality, but not both simultaneously. In our tests, the highest attribution accuracy reached by this variant is 89\%, with a Clip-score of 0.68 and FID of 63.48 (detailed in~\cref{fig:benefits_only_decoder}). The inherent trade-off observed here further reinforces the challenge of balancing these two critical parameters in the context of model fingerprinting techniques.

\begin{figure*}[t]
\centering
  \includegraphics[width=\textwidth]{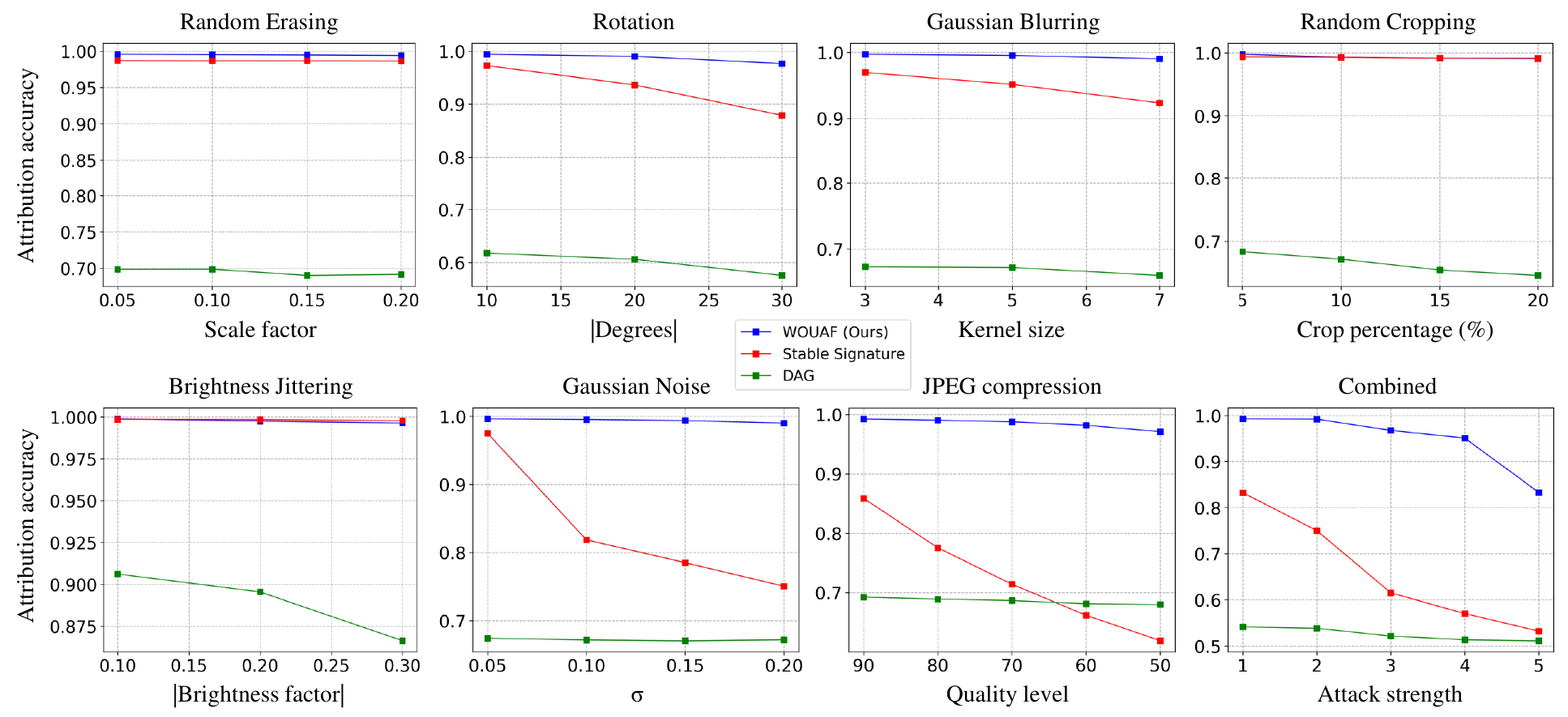}
  \caption{Enhanced Robustness Against Image Post-Processes. For almost all scenarios, WOUAF consistently exceeds the performance of DAG~\cite{kim2020decentralized} and Stable Signature~\cite{stable_signature}.}
  \label{fig:robustness}
 \vspace{-0.15in}
\end{figure*}

\subsection{Robust User Attribution against Image Post-processes} \label{subsec:robustness-postprocesses}
This section evaluates the robustness of our method in scenarios where generated images undergo post-processing. These processes could potentially alter the embedded fingerprint within the images.

Consistent with methodologies outlined in previous research~\citep{kim2020decentralized, yu2020responsible, yu2021artificial, nie2023attributing, stable_signature}, we examine our model’s resilience to various image post-processing operations. We simulate the effect of post-processing at random intensities before inputting data into the fingerprint decoding network, $\mathcal{F}$:
\begin{align}\label{eq:robust}
   L_{\text{robust}} = \mathbb{E}_{z=\mathcal{E}(x), \phi \sim \Phi}\sum_{i=1}^{d_\phi}[\phi_i & \log \sigma(\mathcal{F}(T(\mathcal{D}(\phi, z)))_i \nonumber \\
  +  (1-\phi_i)\log(1-&\sigma(\mathcal{F}(T(\mathcal{D}(\phi, z)))_i], 
\end{align}
where $T(\cdot):\mathbb{R}^{d_x} \rightarrow \mathbb{R}^{d_x}$ denotes the post-processing function. In the optimization process, we employ an objective function akin to the one detailed in~\cref{eq:final_loss}, with $L_\phi$ substituted by $L_{\text{robust}}$.

In our exploration, we contemplate eight different post-processing techniques: Erasing, Rotation, Gaussian Blurring, Cropping, Brightness jittering, the addition of Gaussian Noise, JPEG compression, and a Combination of all these post-processes.
The parameters for these post-processes are designed as follows: For random erasing, we use a random erase ratio within the range [5\%, 10\%, 15\%, 20\%]. Rotation involves randomly sampling a degree within the range (-30,  30). For Gaussian Blurring, we randomly select a kernel size from [3, 5, 7]. For Cropping, we use a random cropping-out ratio within the range [5\%, 10\%, 15\%, 20\%]. The Brightness factor is randomly sampled within the range (-0.3, 0.3). For Gaussian Noise, we add noise with a standard deviation randomly sampled from a uniform distribution $U$[0, 0.2]. JPEG compression quality level is selected from [90, 80, 70, 60, 50]. The Combination technique randomly selects a subset of these seven post-processing methods with a probability of 0.5.

User attribution accuracy for each post-process is evaluated under these parameters. Our tests, depicted in Fig.\ref{fig:robustness}, offer a comparative analysis of user attribution accuracy across robust versions of DAG~\cite{kim2020decentralized}, Stable Signature~\cite{stable_signature}, and WOUAF. Remarkably, our method demonstrates robustness across a range of post-processes, achieving an attribution accuracy improvement of 11\% over Stable Signature and 29\% over DAG. 
A notable trend across all transformations is the monotonic decrease in user attribution accuracy as the intensity of post-processing increases. This reinforces the challenges posed by post-processing in maintaining accurate user attribution. However, our results also underscore the benefits of robust training in overcoming these challenges, emphasizing the importance of resilient training strategies for fingerprinting methods in the face of post-processing transformations.
Considering the robustness of our method against various post-processes, it becomes a viable choice for model distributors seeking reliable fingerprinting solutions. Detailed results of FID scores and visual examples are available in the Appendix.

\section{Deliberate Fingerprint Manipulations} \label{sec:active_adversary}
This section delves into our method's robustness against deliberate attempts to remove fingerprints, which include malicious manipulations via auto-encoders and model purification. Further details and extended attack scenarios are provided in the Appendix.

\subsection{Resilience Against Deep Classifier} \label{sec:secrecy}

The imperceptibility of the fingerprint in generated images is crucial to prevent its detection and subsequent tampering by malicious entities. To assess the secrecy of our method, we adopt an attack scenario akin to the one in~\citep{yu2020responsible}, assuming an attacker aims to train a classifier to detect the presence of a fingerprint.

We assume that the attacker seeks to train a classifier capable of detecting the presence of a fingerprint.
To assess this scenario, we utilize a pretrained ResNet-50~\citep{resnet} based binary classifier, trained using 10K SD generated images (5K original SD images and 5K fingerprinted SD images). This configuration is deemed valid as detecting the presence of a fingerprint necessitates using both non-fingerprinted and fingerprinted images in the training set. The binary classifier achieve 98\% accuracy in the training stage.
In subsequent evaluations using a separate set of 5K images from our variant models, the binary classification accuracy is 0.66 for WOUAF-conv and just 0.56 for WOUAF-all, which is nearly equivalent to \textit{random chance}.

These findings imply that detecting our embedded fingerprint, particularly in the WOUAF-all variant, poses a challenge to detect. Upcoming subsections will delve into further evaluations, predicated on the stringent assumption that users are cognizant of the fingerprint's presence and endeavor to eliminate it by employing auto-encoder methods or fine-tuning techniques.

\subsection{Resilience Against Auto-Encoders} \label{subsec:vae_attack}
In contexts where adversaries aim to alter output images, leveraging deep learning techniques such as neural auto-encoders~\cite{ballemshj18,cheng2020image, minnenbt18} becomes a common strategy for the purpose of obfuscating or removing fingerprints embedded in images~\cite{stable_signature}.
To assess the resilience of our approach, we utilize the robust model against JPEG described in~\cref{subsec:robustness-postprocesses}. This comparison is appropriate as JPEG represents a conventional image compression method. However, the auto-encoders~\cite{ballemshj18,cheng2020image,minnenbt18} employed in our evaluation exhibit superior compression performance compared to JPEG. Our research explores the resilience of our proposed method against these sophisticated auto-encoders, focusing particularly on the impact of their varying compression rates.

\begin{figure}[t]
  \centering
   \includegraphics[width=\linewidth]{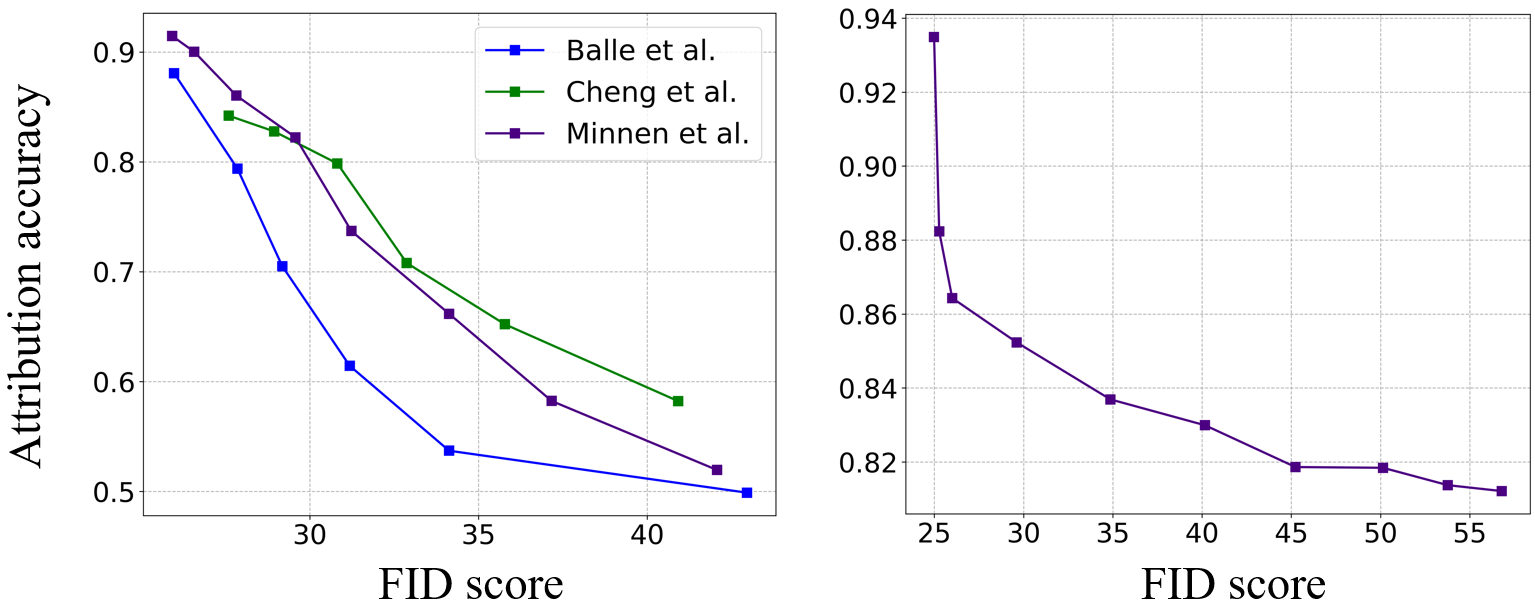}
   \caption{Left: Auto-Encoder-based Fingerprint Removal. With heightened compression rates, both image quality and attribution accuracy experience a decrease. Right: Model Purification. Progressive fine-tuning leads to concurrent declines in both image quality and attribution accuracy. Note that a lower FID score is preferable, indicating better image quality.}
   \label{fig:vae-attack}
   \vspace{-0.15in}
\end{figure}

As depicted on the left side of~\cref{fig:vae-attack}, our investigations reveal a notable trend: attribution accuracy progressively declines towards a near-random level (approximately 50\%) as the compression rate employed by the auto-encoders escalates. This trend highlights a critical trade-off: the reduction in attribution accuracy is achievable solely by compromising the quality of the image~\cite{Li2021ASO}. Our findings indicate that compromising the integrity of the image is a necessary consequence to effectively obscure the fingerprinting process.

\subsection{Resilience Against Model Purification} \label{subsec:network-level-attack}
This subsection addresses the scenario where an adversary, upon recognizing the presence of fingerprints within the images generated by the image decoder $\mathcal{D}$, opts to fine-tune $\mathcal{D}$ with the objective of obliterating the embedded fingerprint. This strategy, known as model purification, is a sophisticated approach to altering the model's output to erase traceable imprints~\cite{stable_signature}.

In this adversarial setting, the primary aim is to refine the downloaded fingerprinted model by optimizing the reconstruction error between the adversary's proprietary image dataset and the output from the fingerprinted model. By adhering to the experimental framework outlined in~\cite{stable_signature}, we charted the interplay between FID scores and attribution accuracy, as presented on the right side of~\cref{fig:vae-attack}. Our empirical analysis reveals a significant challenge: efforts to decrease the attribution accuracy lead to a decline in the quality of the generated images. This finding underscores the inherent complexity in fine-tuning processes aimed at model purification, particularly when striving to maintain the visual quality of the output while endeavoring to obscure its traceable characteristics.
\section{Conclusion}
In this study, we have delved into user attribution for Stable Diffusion-based Text-to-Image (T2I) model, employing a weight modulation-based fingerprinting approach. Our method, WOUAF, not only achieves near-perfect accuracy but also preserves the high quality of generated images. A key aspect of WOUAF is its computational efficiency coupled with enhanced robustness against various image post-processing techniques compared to existing baselines. Our results lay a solid groundwork for future exploration into the broader implications and challenges posed by generative models. In future work, we plan to expand and refine our methodology to encompass various data types including text, audio, and video, necessitating tailored adjustments in model fingerprinting techniques.

\section{Acknowledgment}
This work is partially supported by the National Science Foundation under Grant No. 2038666, No. 2101052, and a grant from Meta AI Learning Alliance. The HuggingFace demo of this work is funded by Intel Corporation. The authors also acknowledge Research Computing at Arizona State University for providing HPC resources and support for this work. The views and opinions of the authors expressed herein do not necessarily state or reflect those of the funding agencies and employers.
\vspace{-24pt}
{
    \small
    \bibliographystyle{ieeenat_fullname}
    \bibliography{main}
}
\clearpage
\maketitlesupplementary
\appendix

\section{Additional Related Work}
\label{app:related_work}
\paragraph{Text-to-Image Generative Models}
Recent advancements in vector quantization and diffusion modeling have significantly enhanced text-to-image (T2I) generation, enabling the creation of hyper-realistic images from textual prompts~\citep{nichol2022glide,saharia2022photorealistic,ramesh2022hierarchical, stable_diffusion}. These T2I models have been effectively utilized in various tasks such as generating images driven by subject, segmentation, and depth cues~\cite{chefer2023attend, chen2023training, patel2023conceptbed, patel2024lambda, gal2022image, li2023gligen}. However, the substantial size of these models presents a challenge for broader user adoption. Research efforts are focusing on enhancing model efficiency through knowledge distillation, step distillation, architectural optimization, and refining text-to-image priors~\cite{li2023snapfusion, salimans2021progressive, luo2023latent,patel2023eclipse}. Amidst these technological advancements, ensuring the responsible usage of these powerful tools is a critical area of focus, which is the aim of our proposed method.

\paragraph{Image Watermarking}
Image watermarking aims to embed a watermark into images for asserting copyright ownership. To maintain the original image's fidelity, these watermarks are embedded imperceptibly. Traditional approaches often employ Fourier or Wavelet transforms, while recent advancements leverage deep neural network-based auto-encoders for this purpose~\cite{rivagan,zhu2018hidden,tancik2020stegastamp}. However, as discussed in the main paper, these methods can be easily disabled in an open-source setting.

From the standpoint of ownership verification, the fingerprinting of generative models aligns conceptually with watermarking techniques. However, unlike direct image manipulation to embed an identifiable signal in watermarking, generative model fingerprinting embeds this signal within the model's weights. Consequently, the identifiable signal is integrated during the image generation process, akin to leaving fingerprints. This approach inherently prevents users from dissociating the fingerprinting process from image generation.

\paragraph{Neural Network Watermarking}
Watermarking techniques, particularly those embedding unique identifiers within model parameters, have been substantively explored in various studies, such as those highlighted in~\cite{adi2018turning, ong2021protecting, uchida2017embedding, darvish2019deepsigns, wang2021riga}. Our methodology, while aligning with the foundational principles of these works, introduces notable advancements in several key areas: utility, scalability, and verification methodology. The majority of existing watermarking techniques are tailored towards image classification models, with only a limited subset extending their applicability to generative models, each presenting its own set of limitations. Unlike traditional methods that predominantly target single classification models, our approach endeavors to fingerprint approximately 4 billion Text-to-Image generator instances through a singular fine-tuning process. Additionally, while prior works have embedded fingerprints into various model aspects, such as input-output dynamics~\cite{adi2018turning, ong2021protecting} or directly within model weights~\cite{uchida2017embedding, darvish2019deepsigns,wang2021riga}, our strategy diverges by eliminating the necessity for trigger input, thereby enhancing scalability. In the context of our problem domain, where malicious users rarely share their model weights with the distributor responsible for watermark verification, the distributor typically only has access to potentially misused images. In essence, our approach not only aligns with but also extends beyond the conventional boundaries of network watermarking techniques, ensuring a thorough inclusion and discussion of these foundational methods in our related works section.

\section{Additional Details}
WOUAF is evaluated utilizing the Stable Diffusion (SD) model~\cite{stable_diffusion} (version 2-base), trained specifically for generating images of 512p resolution.

\section{Additional Experimental Results}
In addition to the figure in the main paper, we added uncurated images using text-prompt from MS-COCO~\cite{lin2014microsoft} and LAION Aesthetics~\cite{schuhmann2022laion}.
For convenience, we have aligned the subsection names with those in the main manuscript. Unless otherwise specified, all figures were generated using the `WOUAF-all' method.

\subsection{Additional Training Details}
The dimension of the mapping network $d_M$ is set to be equal to $4*d_\phi$ across all experimental setups. Training is performed over 50K iterations with a batch size of 32 and a learning rate of $10^{-4}$ using AdamW optimizer~\cite{adamw}. 

\subsection{Attribution Accuracy and Image Quality}

\begin{figure*}[!htbp]
\centering
  \includegraphics[width=\textwidth]{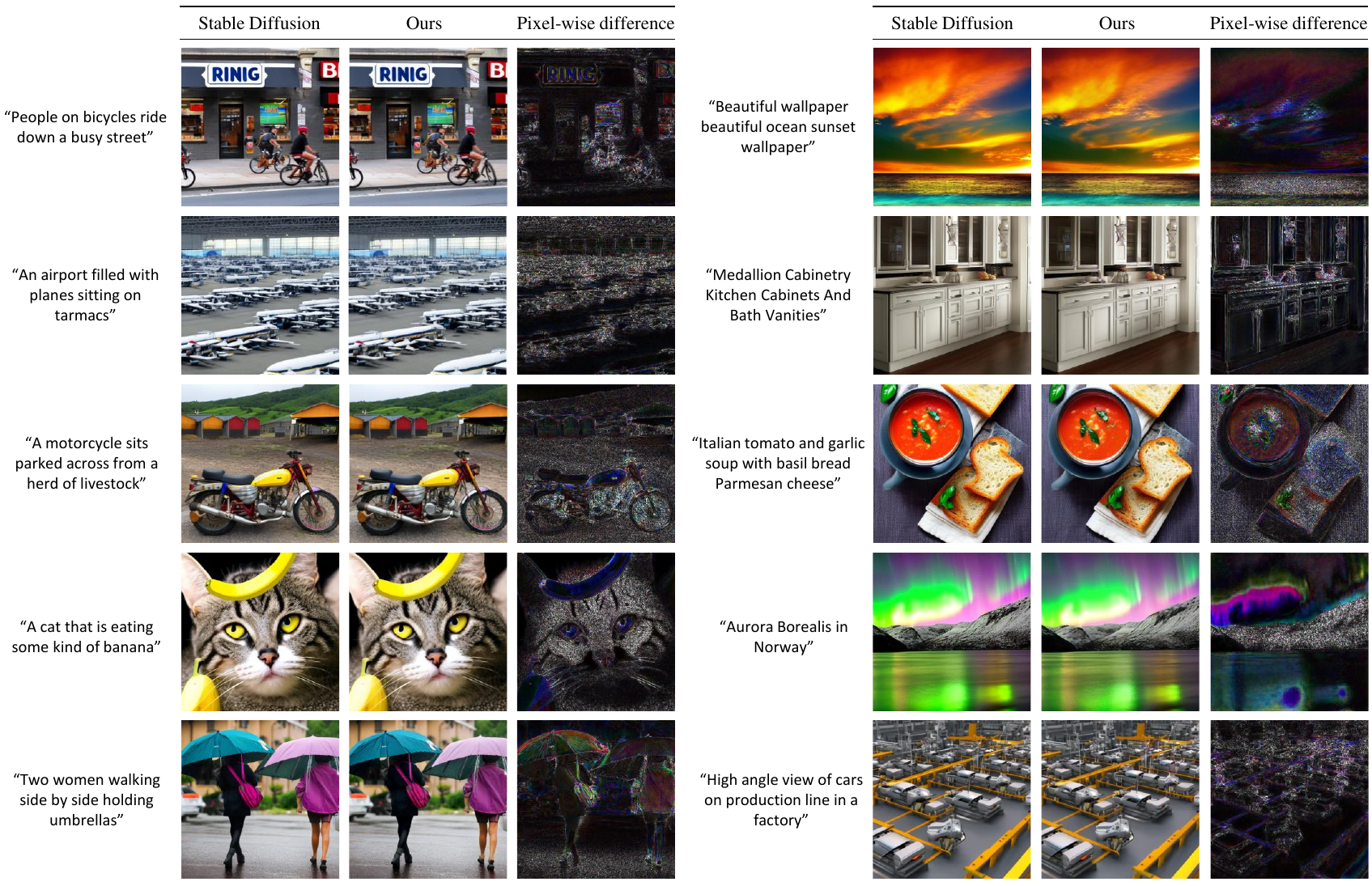} \\ \hspace*{6.5em}(a) MS COCO \hspace{12.5em}(b) LAION Aesthetics
  \caption{Uncurated images of the original and fingerprinted Stable Diffusion models on MS-COCO and LAION Aesthetics. Pixel-wise differences are multiplied by a factor of 5 for a better view.}
  \label{fig:supp_uncurated}
\end{figure*}

As highlighted in the main manuscript, our methodology has a negligible effect on the original Stable Diffusion's image quality. Please refer to Fig.~\ref{fig:supp_uncurated} for these uncurated images.

\subsection{Evaluating Generalizability Across Datasets} 
A key feature of our proposed methodology is its design independence from image-text paired datasets for achieving attribution accuracy. This property imbues it with the potential for broad applicability across a diverse range of contexts. To substantiate this claim, we conducted an experiment in which our variant models were trained exclusively on the ImageNet dataset~\cite{deng2009imagenet}.
We subsequently evaluated the performance of these ImageNet-trained models on the MS-COCO test set as well as a randomly selected portion of the LAION-aesthetics datasets. 

\begin{figure*}[!htbp]
\centering
  \includegraphics[width=\textwidth]{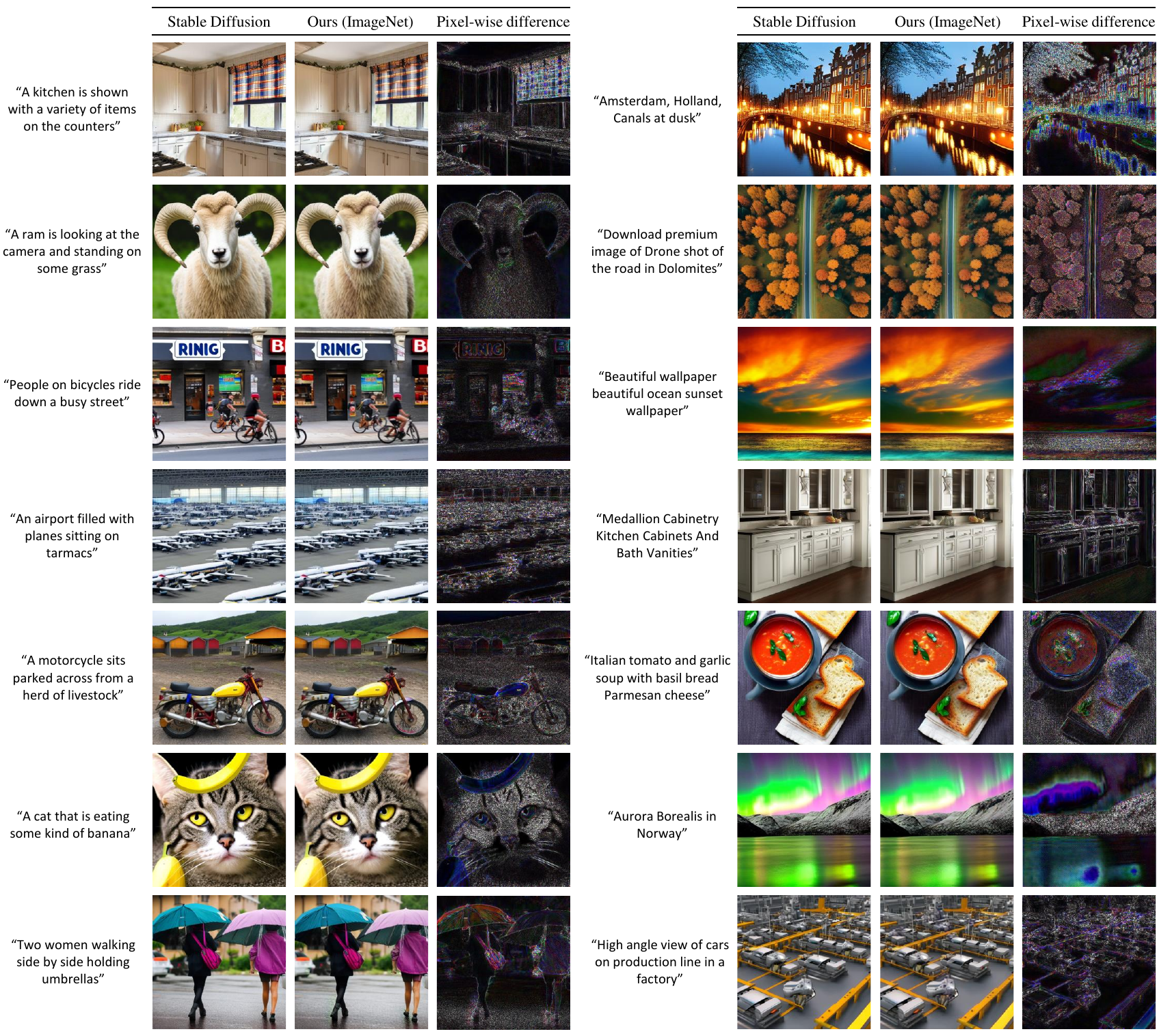} \\ \hspace*{6.5em}(a) MS COCO \hspace{12.5em}(b) LAION Aesthetics
  \caption{Qualitative comparisons of the original and fingerprinted Stable Diffusion models that were fine-tuned using only the ImageNet dataset. Pixel-wise differences are multiplied by a factor of 5 for a better view.}
  \label{fig:supp_imagenet}
\end{figure*}

The evaluation results, as seen in Table~\ref{tab:attribution_imagenet}, effectively corroborate our assertion. Our methodology demonstrates compelling performance, with both our variants, achieving high attribution accuracy and maintaining image generation quality. These results underscore our method's independence from the use of text-image paired datasets, thereby establishing its broad applicability in diverse scenarios where reliable fingerprinting and high-quality image generation are required. Fig.~\ref{fig:supp_imagenet} provides a visual representation of these images.

\begin{table*}[!htbp]
\centering
\caption{Assessment of attribution accuracy and generation quality using Imagenet trained models. We validated our method using MS-COCO testset and LAION-aesthetics dataset. $\uparrow$/$\downarrow$ indicates higher/lower is desired.}
\label{tab:attribution_imagenet}
{\small
\begin{tabularx}{\textwidth}{lXXXXXX}
\toprule
\multirow{3}{*}{Model} & \multicolumn{3}{c}{MS-COCO} & \multicolumn{3}{c}{LAION} \\
 & Attribution Acc. ($\uparrow$) & Clip-score ($\uparrow$)& FID ($\downarrow$)& Attribution Acc. ($\uparrow$) & Clip-score ($\uparrow$)& FID ($\downarrow$)\\
\midrule
Ours-conv. & 0.99 & 0.73 & 24.23 & 0.99 & 0.51 & 19.71 \\
Ours-all & 0.99 & 0.73 & 24.41 & 0.99 & 0.51 & 19.46 \\
\bottomrule
\end{tabularx}
}
\vspace{-0.1in}
\end{table*}

\subsection{Attribution Accuracy Across Various Generation Hyperparameters}

In accordance with the details provided in the primary manuscript, we subjected our methodology to evaluation employing two widely accepted schedulers: Euler~\cite{euler}, featuring time steps at intervals of [15, 20, 25], and DDIM~\cite{ddim}, operating at time steps in [45, 50, 55]. Along with these, we also incorporated classifier-free guidance scales~\cite{ho2022classifier} at 2.5, 5.0, and 7.5.

\begin{figure*}[t]
\centering
  \includegraphics[width=\textwidth]{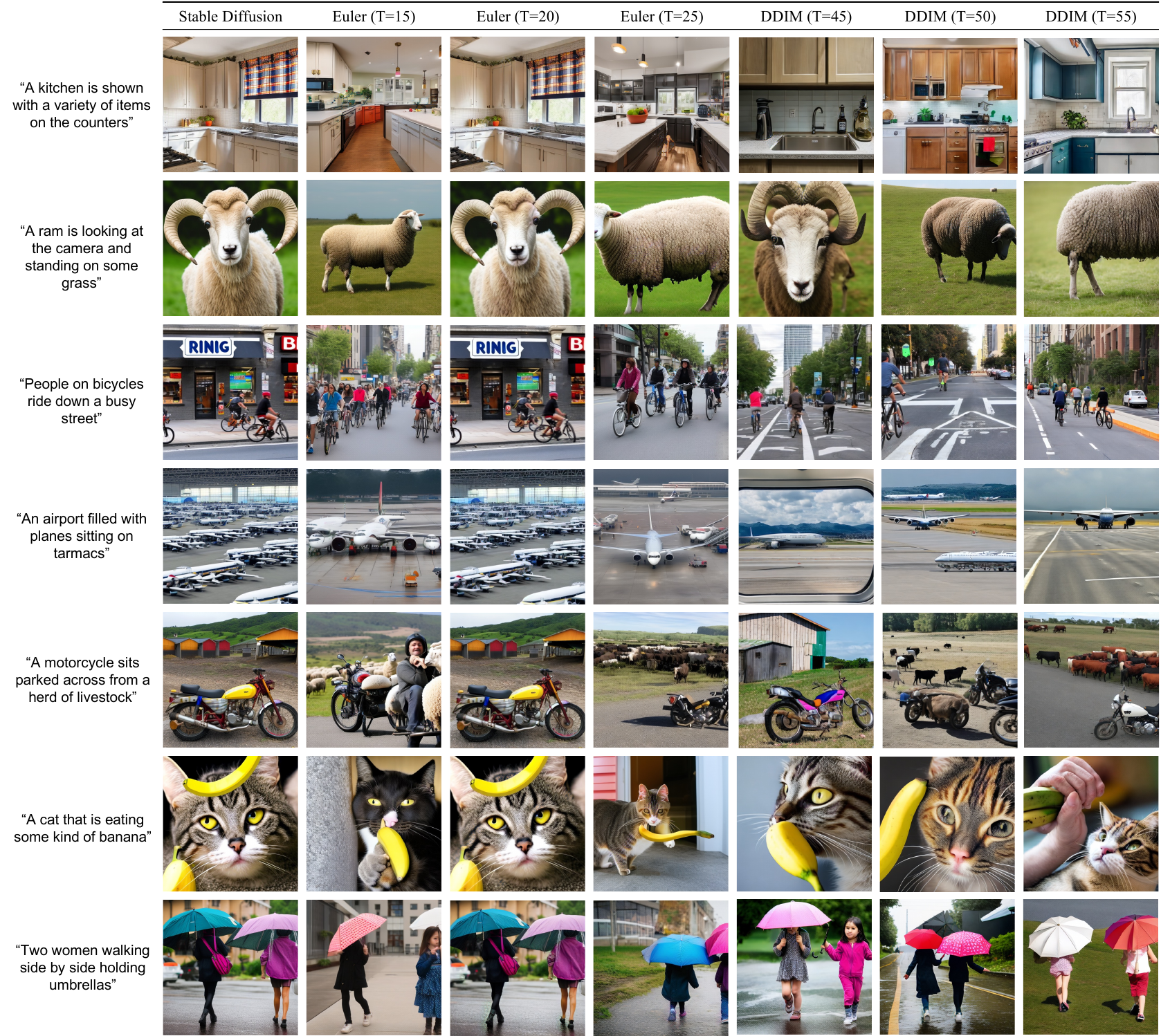}
  \caption{Qualitative results obtained using the Euler and DDIM schedulers with varying time steps on the MS-COCO dataset. We maintained a constant classifier-free guidance scale~\cite{ho2022classifier} at \textbf{7.5}. Each column corresponds to the 'WOUAF-all' rows in Table~\ref{tab:hyperparam_1}.}
  \label{fig:supp_various_1}
\end{figure*}

\begin{figure*}[t]
\centering
  \includegraphics[width=0.88\textwidth]{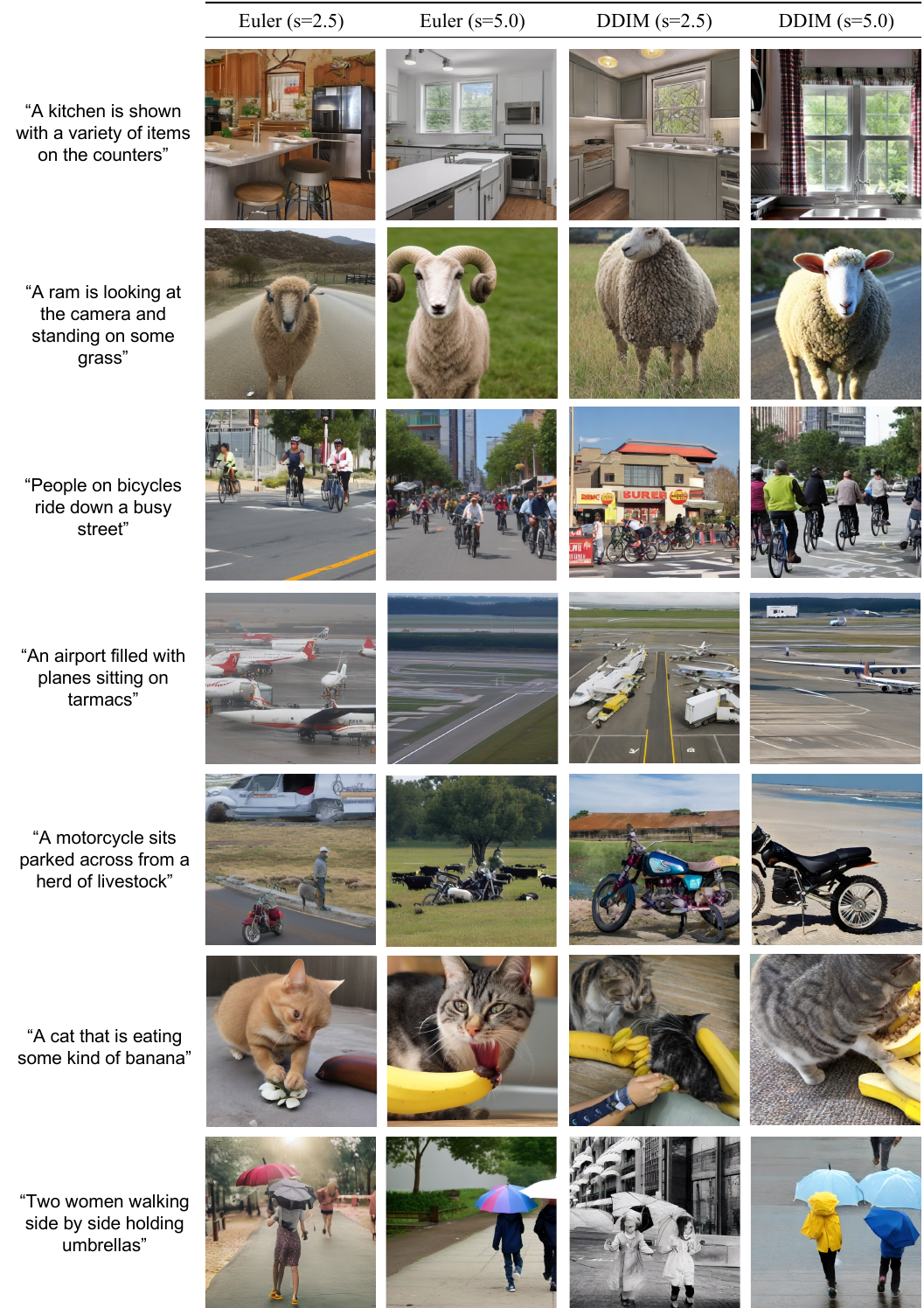}
  \caption{Qualitative results produced by applying different classifier-free guidance scales~\cite{ho2022classifier} on the MS-COCO dataset. The scheduler and time steps were held constant at Euler for \textbf{20} steps and DDIM for \textbf{50} steps. Each column aligns with the 'WOUAF-all' rows in Table~\ref{tab:hyperparam_2}.}
  \label{fig:supp_various_2}
\end{figure*}

Echoing the discussions in the main paper, the data in Tab.~\ref{tab:hyperparam_1} and~\ref{tab:hyperparam_2} corroborate the near-perfect attribution accuracy achieved by our method. Furthermore, the absence of significant deterioration in quality metrics reaffirms the resilience of our approach in the face of diverse generation hyperparameters (Refer to Fig.~\ref{fig:supp_various_1} and Fig.~\ref{fig:supp_various_2}).

\begin{table*}[t]
\centering
\caption{Assessment of attribution accuracy and generation quality using Euler and DDIM scheduler with different time steps on MS-COCO. We fixed classifier-free guidance scale~\cite{ho2022classifier} to \textbf{7.5}.
$\uparrow$/$\downarrow$ indicates higher/lower is desired.}
\label{tab:hyperparam_1}
{\small
\begin{tabularx}{\textwidth}{lXXXXXXXXX}
\toprule
\multirow{3}{*}{Model} & \multicolumn{4}{c}{Euler~\cite{euler}} & \multicolumn{4}{c}{DDIM~\cite{ddim}} \\\cmidrule(lr){2-5} \cmidrule(lr){6-9}
  & Steps & Attribution Acc. ($\uparrow$)  & CLIP-score ($\uparrow$)& FID ($\downarrow$) & Steps & Attribution Acc. ($\uparrow$) & CLIP-score ($\uparrow$)& FID ($\downarrow$)\\
\midrule
Original SD~\cite{stable_diffusion} & 20 & - & 0.73 & 24.48 & 50 & - & 0.73 & 23.33 \\
\midrule
\multirow{3}{*}{WOUAF-conv} & 15 & 0.99 & 0.73 & 24.63 & 45 & 0.99 & 0.73 & 23.28 \\
 & 20 & 0.99 & 0.73 & 24.43 & 50 & 0.99 & 0.73 & 23.35 \\
 & 25 & 0.99 & 0.74 & 24.14 & 55 & 0.99 & 0.73 & 23.31 \\
\midrule
\multirow{3}{*}{WOUAF-all} & 15 & 0.99 & 0.73 & 24.65 & 45 & 0.99 & 0.73 & 23.34 \\
 & 20 & 0.99 & 0.73 & 24.42 & 50 & 0.99 & 0.73 & 23.29 \\
 & 25 & 0.99 & 0.73 & 24.11 & 55 & 0.99 & 0.73 & 23.26 \\
\bottomrule
\end{tabularx}
}
\vspace{-0.1in}
\end{table*}

\begin{table*}[t]
\centering
\caption{Assessment of attribution accuracy and generation quality on different classifier-free guidance scales~\cite{ho2022classifier} using MS-COCO. We fixed the scheduler and time steps to Euler for \textbf{20} steps and DDIM for \textbf{50} steps.
$\uparrow$/$\downarrow$ indicates higher/lower is desired.}
\label{tab:hyperparam_2}
{\small
\begin{tabularx}{\textwidth}{lXXXXXXXXX}
\toprule
\multirow{3}{*}{Model} & \multicolumn{4}{c}{Guidance Scale 2.5} & \multicolumn{4}{c}{Guidance Scale 5.0} \\\cmidrule(lr){2-5} \cmidrule(lr){6-9}
  & Scheduler & Attribution Acc. ($\uparrow$)  & CLIP-score ($\uparrow$)& FID ($\downarrow$) & Scheduler & Attribution Acc. ($\uparrow$) & CLIP-score ($\uparrow$)& FID ($\downarrow$)\\
\midrule
\multirow{2}{*}{WOUAF-conv} & Euler & 0.99 & 0.72 & 18.63 & Euler & 0.99 & 0.73 & 21.91 \\
 & DDIM & 0.99 & 0.72 & 18.35 & DDIM & 0.99 & 0.73 & 20.78 \\
\midrule
\multirow{2}{*}{WOUAF-all} & Euler & 0.99 & 0.71 & 18.64 & Euler & 0.99 & 0.73 & 21.89 \\
 & DDIM & 0.99 & 0.71 & 18.31 & DDIM & 0.99 & 0.73 & 20.68 \\
\bottomrule
\end{tabularx}
}
\vspace{-0.1in}
\end{table*}

\subsection{Benefits of Finetuning only Decoder}

\begin{figure*}[t]
\centering
  \includegraphics[width=\textwidth]{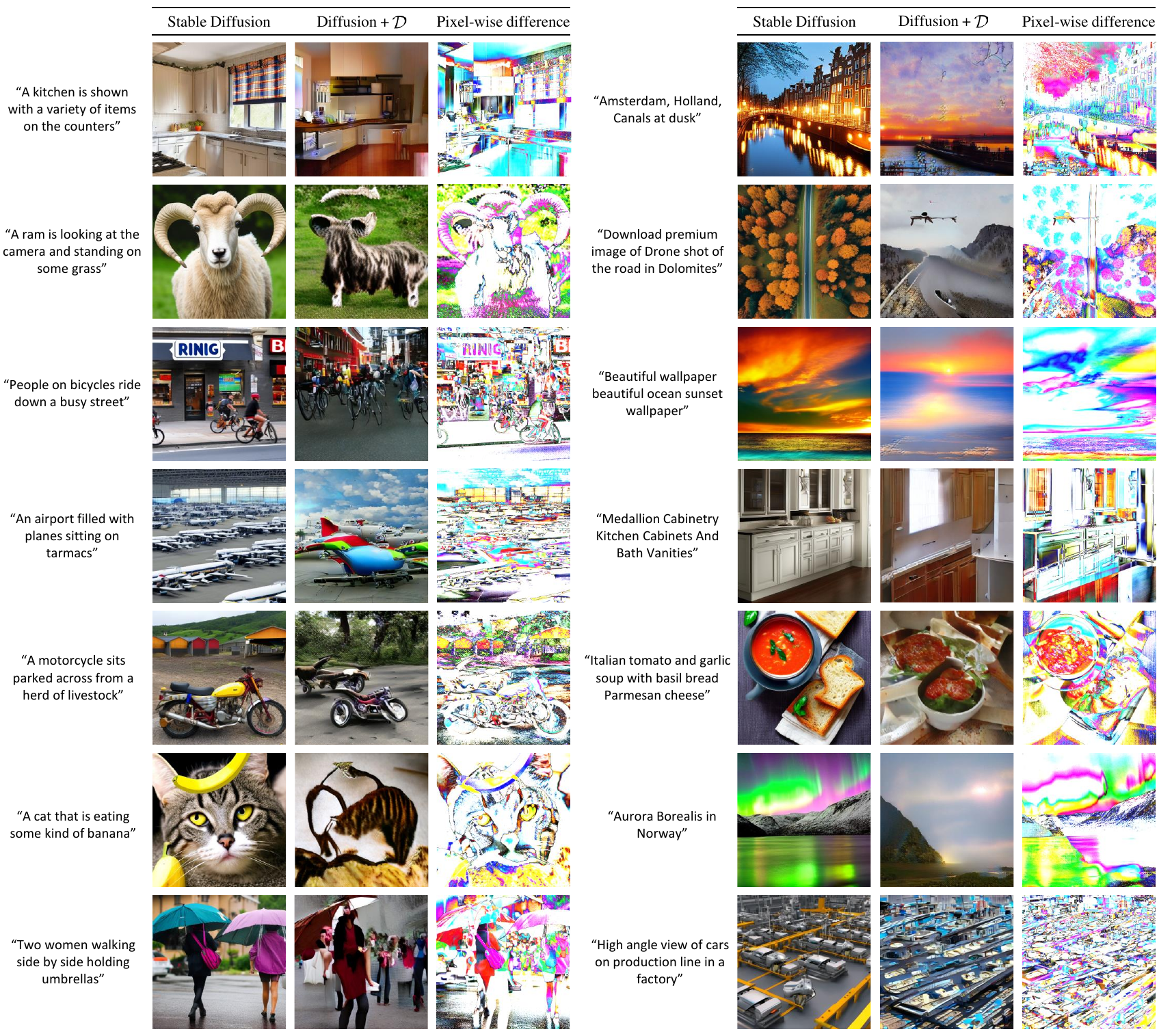} \\ \hspace*{6.5em}(a) MS COCO \hspace{12.5em}(b) LAION Aesthetics
  \caption{Qualitative results of the original and fingerprinted Stable Diffusion models on MS-COCO and LAION Aesthetics. When fine-tuning the SD model's $\epsilon_\theta$ and $\mathcal{D}$ together, there are significant quality drops. Pixel-wise differences are multiplied by a factor of 5 for a better view.}
  \label{fig:supp_unet_decoder}
\end{figure*}

In this section, we present qualitative outcomes resulting from the joint fine-tuning of the Stable Diffusion model's components, diffusion model $\epsilon_\theta$ and decoder $\mathcal{D}$. 
As accentuated in the primary manuscript, our training protocol achieved an accuracy of 89\%, however, it resulted in a noticeable deterioration in the quality metrics (Clip-score: 0.68, FID: 63.48).
Fig.~\ref{fig:supp_unet_decoder} provides additional visual affirmation of these quantitative results.

\subsection{Robust User Attribution against Image Post-processes} \label{sup:robust_image_post_processes}

\begin{figure*}[ht]
\centering
  \includegraphics[width=\textwidth]{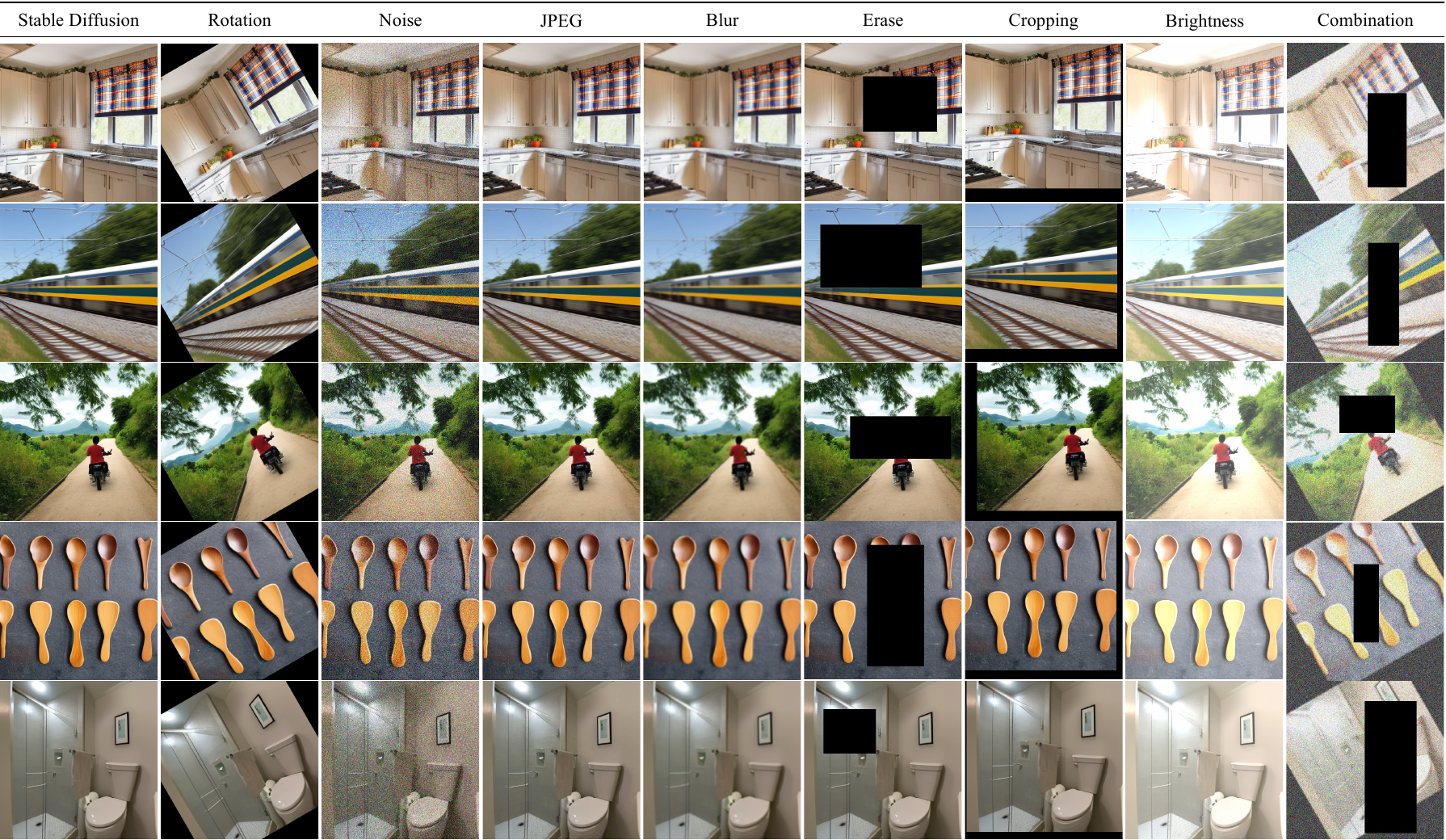} \\ 
  \caption{Qualitative results following the application of post-processing: These images demonstrate that intensive post-processing significantly compromises the perceptual quality of images, thereby deterring both malicious and benign users from employing robust post-processing techniques.}
  \label{fig:atk_samples}
\end{figure*}

\begin{figure*}[t]
\centering
  \includegraphics[width=\textwidth]{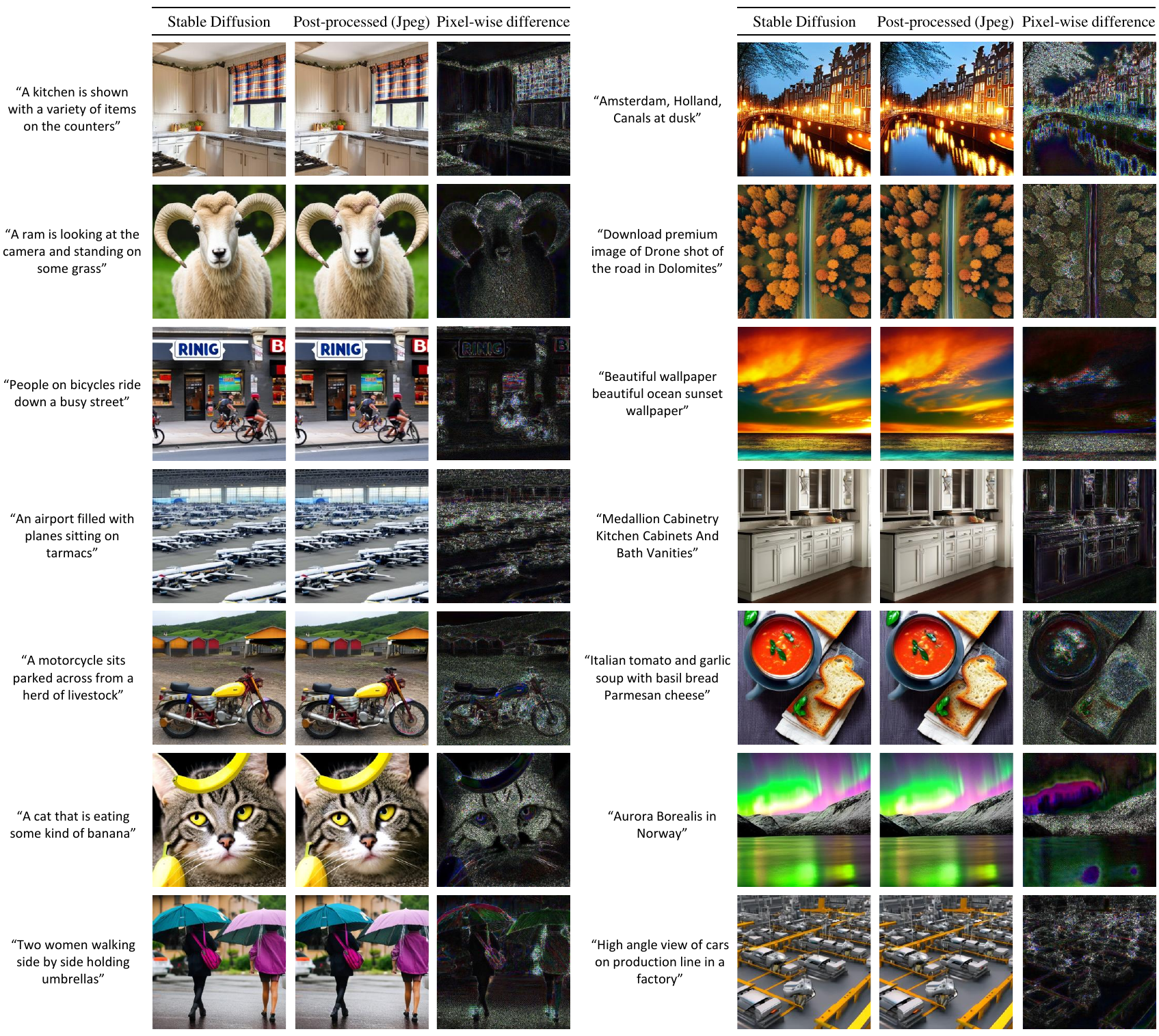} \\ \hspace*{6.5em}(a) MS COCO \hspace{12.5em}(b) LAION Aesthetics
  \caption{Qualitative results of the original and fingerprinted Stable Diffusion models on MS-COCO and LAION aesthetics. Our fingerprinted model is trained by simulating JPEG compression during training. Pixel-wise differences are multiplied by a factor of 5 for a better view.}
  \label{fig:supp_post_jpeg}
\end{figure*}

\begin{figure*}[t]
\centering
  \includegraphics[width=\textwidth]{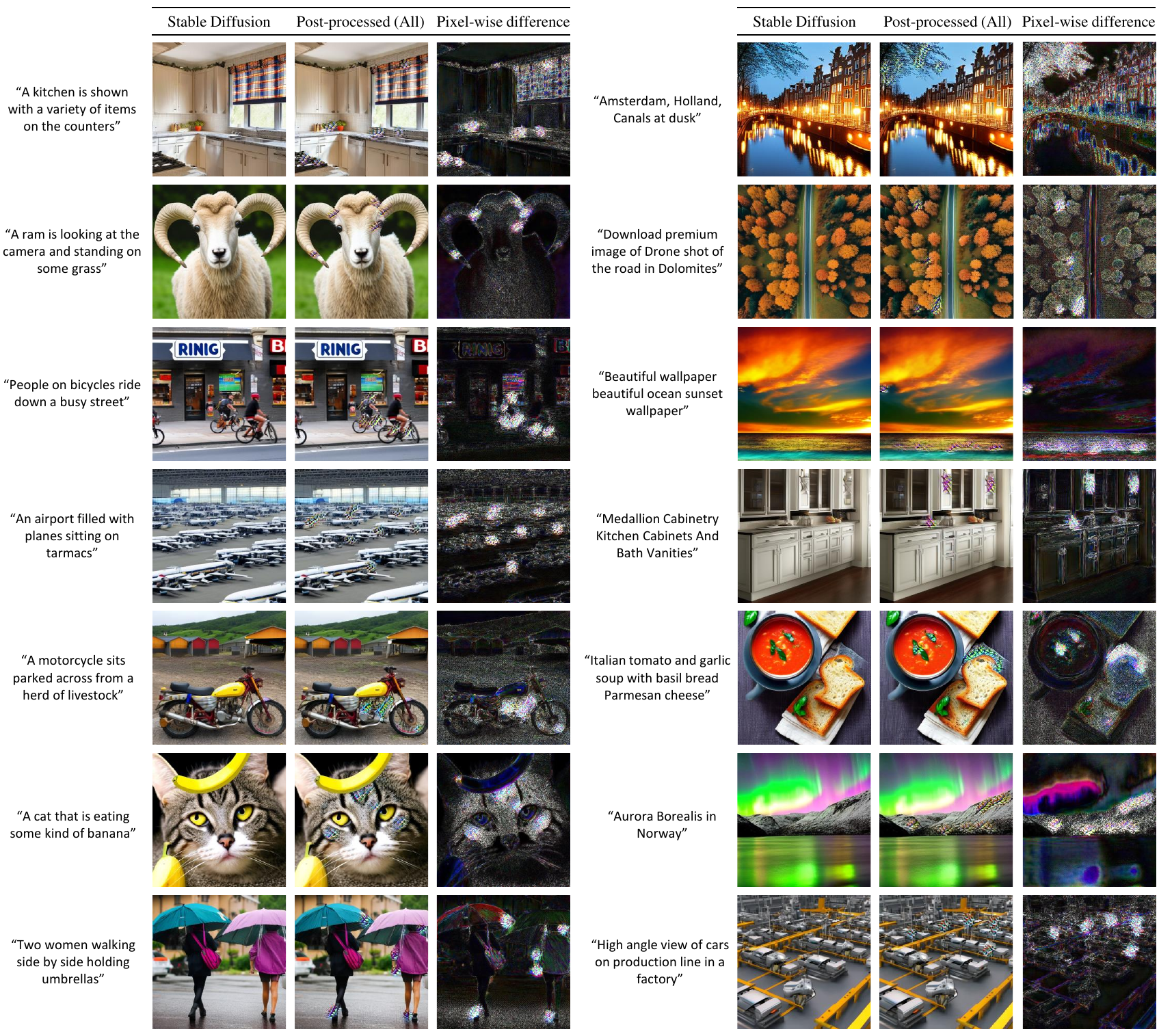} \\ \hspace*{6.5em}(a) MS COCO \hspace{12.5em}(b) LAION Aesthetics
  \caption{Qualitative results of the original and fingerprinted Stable Diffusion models on MS-COCO and LAION aesthetics. Our fingerprinted model is trained by simulating all the combinations of the post-processing during training. Pixel-wise differences are multiplied by a factor of 5 for a better view.}
  \label{fig:supp_post_all}
\end{figure*}

We conducted a thorough evaluation of quality metrics to assess the impact of our robust user attribution training on various image post-processing methods. Examples of images post-processed using these methods are displayed in~\cref{fig:atk_samples}. As indicated in~\cref{tab:robust_fid} and~\cref{tab:robust_clip}, our robust fine-tuning approach generally preserves image quality with only minimal perturbations. A representative example under a JPEG attack, generated by our robust model, is showcased in Fig.~\ref{fig:supp_post_jpeg}.
Additionally, our method demonstrates adaptability under Combination attacks, which significantly challenge image fidelity. As illustrated in~\cref{fig:atk_samples}, these combined post-processing techniques necessitate a relatively stronger fingerprint compared to single post-processes, as further detailed in~\cref{fig:supp_post_all}. Moreover, it is observed that images subjected to extensive post-processing lose perceptual value, impacting both malicious and naive users alike.

\subsection{Evaluating Robustness Beyond Training Configurations}

\begin{table*}[!ht]
\centering
\caption{Assessment of generalizability of robust WOUAF. We measure the attribution accuracy for different models.}
\label{tab:generalizability}
{\small
\begin{tabularx}{\textwidth}{lXXXXXXX}
\toprule
& Crop& Rotate& Blur & Bright& Noise & Erase \\
Params  & 30\% & $\pm$ 40& 9 & $\pm$ 0.4& 0.3 & 30\% \\
\midrule
Stable Signature~\cite{stable_signature} & 0.997 & 0.809 & 0.873 & 0.996 & 0.737 & 0.986 \\
WOUAF-all & 0.946 & 0.923 & 0.657 & 0.975 & 0.970 & 0.991 \\
\bottomrule
\end{tabularx}
}
\vspace{-0.1in}
\end{table*}

In~\cref{subsec:robustness-postprocesses} and~\cref{sup:robust_image_post_processes}, we explore the resilience of our methodology against a variety of image post-processing techniques. Our experiments are designed with predefined post-process strengths sufficient to deter both malicious and benign uses of strong post-process modifications (refer to~\cref{fig:atk_samples}). Nonetheless, it is imperative to evaluate the resilience of our approach and the established baselines under conditions involving more intense attacks than those encountered during the training phase.
In~\cref{tab:generalizability}, we execute a series of tests to measure the attribution accuracy under more severe perturbations than those used in training and compare these results against those of baseline methodologies.

\begin{table*}[t]
\centering
\caption{FID~\cite{heusel2017gans} scores using MS-COCO after robust training. 
Lower is desired.}
\label{tab:robust_fid}
{\small
\begin{tabularx}{\textwidth}{lXXXXXXXXX}
\toprule
Model & Crop & Rotation & Blur& Brightness& Noise & Erasing & JPEG& Combi.\\
\midrule
WOUAF-conv & \multirow{2}{*}{24.02} & \multirow{2}{*}{24.05} & \multirow{2}{*}{23.80} & \multirow{2}{*}{24.14} & \multirow{2}{*}{23.96} & \multirow{2}{*}{24.16} & \multirow{2}{*}{24.42} & \multirow{2}{*}{26.80} \\
(robust) &  &  &  &  &  &  &  &  \\
WOUAF-all & \multirow{2}{*}{24.35} & \multirow{2}{*}{23.92} & \multirow{2}{*}{24.18} & \multirow{2}{*}{24.54} & \multirow{2}{*}{24.24} & \multirow{2}{*}{24.48} & \multirow{2}{*}{24.41} & \multirow{2}{*}{26.85} \\
(robust) &  &  &  &  &  &  &  &  \\
\bottomrule
\end{tabularx}
}
\vspace{-0.1in}
\end{table*}

\begin{table*}[t]
\centering
\caption{CLIP scores~\cite{hessel2021clipscore} using MS-COCO after robust training. 
Higher is desired.}
\label{tab:robust_clip}
{\small
\begin{tabularx}{\textwidth}{lXXXXXXXXX}
\toprule
Model & Crop & Rotation & Blur& Brightness& Noise & Erasing & JPEG& Combi.\\
\midrule
WOUAF-conv & \multirow{2}{*}{0.716} & \multirow{2}{*}{0.717} & \multirow{2}{*}{0.716} & \multirow{2}{*}{0.716} & \multirow{2}{*}{0.712} & \multirow{2}{*}{0.717} & \multirow{2}{*}{0.714} & \multirow{2}{*}{0.702} \\
(robust) &  &  &  &  &  &  &  &  \\
WOUAF-all & \multirow{2}{*}{0.719} & \multirow{2}{*}{0.717} & \multirow{2}{*}{0.718} & \multirow{2}{*}{0.718} & \multirow{2}{*}{0.710} & \multirow{2}{*}{0.718} & \multirow{2}{*}{0.716} & \multirow{2}{*}{0.704} \\
(robust) &  &  &  &  &  &  &  &  \\
\bottomrule
\end{tabularx}
}
\vspace{-0.1in}
\end{table*}

\begin{figure*}[t]
\centering
  \includegraphics[width=\textwidth]{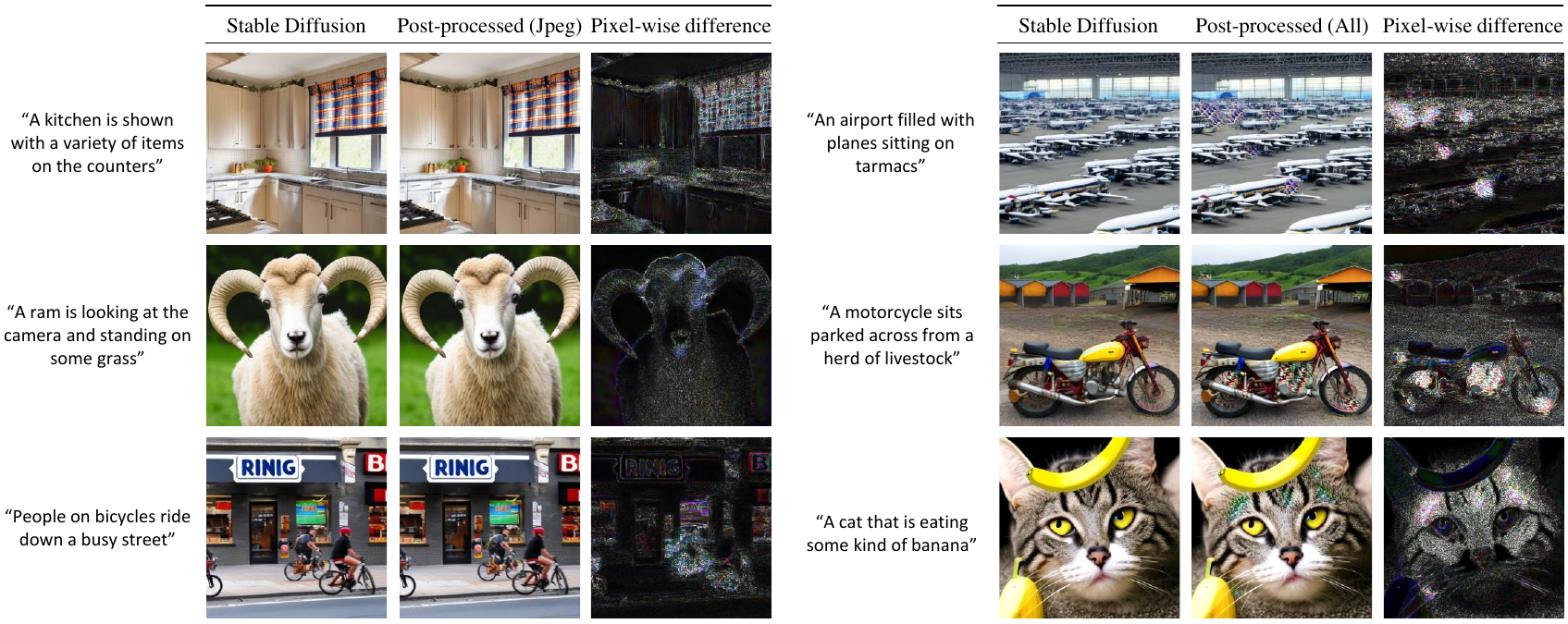}
  \caption{Qualitative results of the original and fingerprinted Stable Diffusion models (WOUAF-conv) on MS-COCO. Pixel-wise differences are multiplied by a factor of 5 for a better view.}
  \label{fig:supp_post_de}
\end{figure*}

\section{Additional Deliberate Fingerprint Manipulation}
\subsection{Gaussian Noise Model Purification}
\begin{figure}[!htbp]
\centering
  \includegraphics[width=\linewidth]{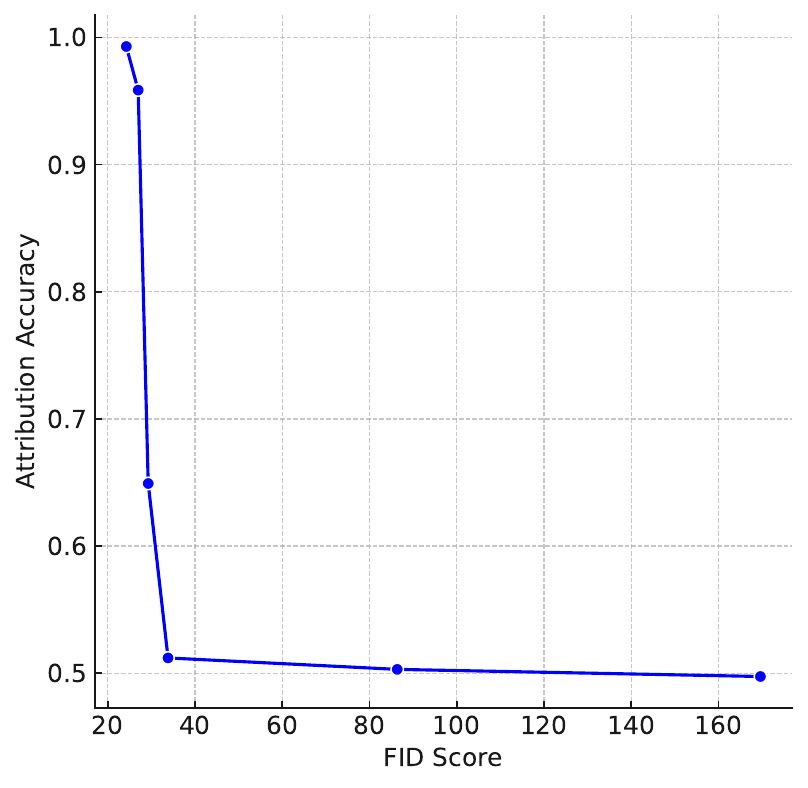}
  \caption{Model Purification. Adding Gaussian noise into weights leads to concurrent declines in both image quality and attribution accuracy. Note that a lower FID score is preferable, indicating better image quality.}
  \label{fig:gaussian_atk}
\end{figure}
This subsection addresses the scenario where an adversary, upon recognizing the presence of fingerprints within the images generated by the image decoder $\mathcal{D}$, opts to add Gaussian noise into $\mathcal{D}$ to obliterate the embedded fingerprint. In order to test this scenario, we gradually increase the standard deviation following $[0., 0.01, 0.015, 0.02, 0.025, 0.03]$. As shown in~\cref{fig:gaussian_atk}, our empirical analysis reveals a significant challenge: efforts to decrease the attribution accuracy lead to a decline in the quality of the generated images. This result also supports the idea that efforts to decrease attribution accuracy lead to a significant decline in the quality of the generated images.

\subsection{Full Knowledge Attack Scenario}
This scenario assumes an internal attacker with comprehensive knowledge of our training process, including the training dataset, model structure, fingerprint space, and training details. To validate this, we trained an attacker's version, following our methodology but employing a different random seed. We then assessed user attribution accuracy by inputting 5K images generated by the attacker's model into WOUAF-conv and WOUAF-all fingerprint decoding networks. Both of our model variants exhibited user attribution accuracies of \textbf{0.509} and \textbf{0.501}, which are essentially random guesses, and thus dodged the attack. Even when an attacker with complete knowledge replicates our methodology, they will not be able to mislead the original fingerprint decoding network.


\end{document}